\documentclass{article}

\PassOptionsToPackage{numbers}{natbib}  
 \usepackage[eandd, final]{neurips_2026}

\usepackage{multirow}
\usepackage[utf8]{inputenc} %
\usepackage[T1]{fontenc}    %
\usepackage[hidelinks]{hyperref}       %
\usepackage{url}            %
\usepackage{booktabs}       %
\usepackage{amsfonts}       %
\usepackage{nicefrac}       %
\usepackage{microtype}      %
\usepackage{xcolor}         %
\newcommand{\rev}[1]{#1}
\usepackage{colortbl}       %

\usepackage{subcaption}
 \usepackage{rotating}
\usepackage{comment}
\usepackage{fvextra}
\newcommand{\repo}{\href{https://github.com/YahyaAalaila/seahorse}{\small\texttt{github.com/YahyaAalaila/seahorse}}}
\newcommand{\docs}{\href{https://yahyaaalaila.github.io/seahorse/}{\small\texttt{yahyaaalaila.github.io/seahorse}}}

\DefineVerbatimEnvironment{codeblock}{Verbatim}{
  breaklines=true,
  breakanywhere=true,
  fontsize=\small,
  frame=single,
  framerule=0pt,
  rulecolor=\color{gray!40},
  framesep=3mm,
  bgcolor=blue!3
}

\usepackage{amsmath} 
\usepackage{graphicx}

\usepackage{tikz}
\usetikzlibrary{arrows.meta, positioning, fit}

\usepackage{amsmath,amsfonts,amssymb}
\DeclareMathAlphabet{\mathbbold}{U}{bbold}{m}{n}

\usepackage{natbib}

\usepackage{tikz}
\usetikzlibrary{positioning, fit, backgrounds, arrows.meta}

\usepackage{booktabs}
\usepackage{tabularx}
\usepackage{array}
\usepackage{enumitem}
\newcolumntype{Y}{>{\raggedright\arraybackslash}X}

\title{\textsc{Seahorse}:\\A Unified Benchmarking Framework for Spatiotemporal Event Modeling}

\author{%
  Yahya Aalaila\\
  DFKI\thanks{DFKI: Deutsches Forschungszentrum für Künstliche Intelligenz GmbH (German Research Center for Artificial Intelligence). 
  Corresponding author: \texttt{Yahya.Aalaila@dfki.de}}, \;RPTU\thanks{Rheinland-Pfälzische Technische Universität Kaiserslautern-Landau (RPTU)} \\
  Kaiserslautern, Germany \\
  {\scriptsize \texttt{Yahya.Aalaila@dfki.de}} \\
  \And
  Sebastian Vollmer \\
  DFKI\footnotemark[1], \;RPTU\footnotemark[2] \\
  Kaiserslautern, Germany \\
  {\scriptsize \texttt{Sebastian.Vollmer@dfki.de}} \\
  \And
  Gerrit Gro{\ss}mann \\
  DFKI\footnotemark[1] \\
  Kaiserslautern, Germany \\
  {\scriptsize \texttt{Gerrit.Grossmann@dfki.de}} \\
}

\begin{document}

\maketitle

\begin{abstract}
Spatiotemporal point processes (STPPs) model event data in continuous time and space, with applications in mobility, epidemiology, and public safety. Recent neural STPPs span expressive intensities, density models, continuous-time latent dynamics, normalizing-flow spatial decoders, and score-based generative mechanisms. Yet comparison remains fragile because implementations differ in preprocessing, coordinate normalization, splits, likelihood conventions, and evaluation protocols. We present \textsc{Seahorse}, a unified framework for reproducible training and evaluation of STPPs. \textsc{Seahorse} provides an encode--evolve--decode interface and an executable benchmarking contract that standardizes dataset handling, tuning, raw-coordinate likelihood reporting, and run artifacts across classical baselines, neural likelihood models, continuous-time Neural STPP variants, and sample-based generative models. On 13 real datasets and 19 presets, no model dominates: likelihood, calibration, and next-event prediction favor different models, and the strongest models take from minutes to hours to train. Using HawkesNest as a controlled synthetic stress-test suite, we show how diagnostic benchmarking can probe neural STPP behavior under increasing spatiotemporal entanglement. \textsc{Seahorse} provides reusable infrastructure for comparing, diagnosing, and extending STPP models beyond implementation-specific leaderboards. The implementation, including preset configurations and reproduction scripts, is available at \repo.
\end{abstract}

\section{Introduction}
\label{dec:introduction}
\rev{Emergency calls, earthquakes and disease cases are recorded as events with a time and a place, and modeling such data means describing when and where events occur. A common approach aggregates events into counts on a space--time grid and applies time-series or deep forecasting models; others keep only the event times (temporal point processes) or only the locations (spatial point patterns). Spatiotemporal point processes (STPPs) instead model each event's time and location jointly, in continuous time and space} \citep{daley2007introduction,reinhart2018review}. Neural point process models have substantially expanded the classical toolkit: recurrent and attention-based encoders model complex event histories \citep{du2016recurrent,mei2017neural,zuo2020transformer}, while recent neural STPPs introduce continuous-time latent dynamics, learned spatial densities, exact-integration networks, and generative mechanisms for events in continuous space and time \citep{chen2021neural,zhou2022neural,zhou2024automatic}. This diversity is scientifically useful, but it makes empirical comparison difficult because model families expose different predictive objects, including intensities, conditional densities, cumulative hazards, and generative samplers, and therefore require different training and evaluation machinery.
\begin{figure}[t]
    \centering
    \includegraphics[width=\linewidth]{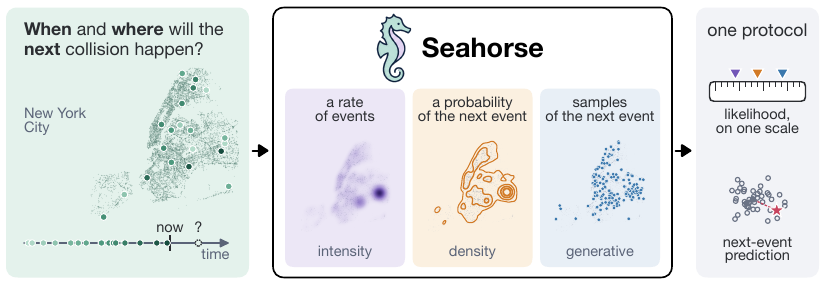}
\caption{
\textbf{Overview of \textsc{Seahorse}.}
\rev{Left: the most recent collisions before ``now'' on one New York City test day. Middle: STPP models answer the same question in different forms, as an event rate, a probability of the next event, or samples of the next event; all three are illustrated with one simple self-exciting model on these events. Right: \textsc{Seahorse} evaluates them under one protocol (schematic).}
}
    \label{fig:seahorse_overview}
\end{figure}

Despite rapid progress, evidence for neural STPPs remains fragmented. Studies differ in preprocessing, spatial normalization, data splits, support construction, coordinate transforms, numerical integration, and likelihood reporting conventions \citep{mukherjee2025neural,cheng2025deep}. This is not merely an engineering reproducibility issue. In STPPs, likelihood values depend on the coordinate measure and support used for density evaluation, while surrogate objectives such as variational bounds, score matching, or imitation losses are not directly comparable to exact point-process likelihoods without explicit evaluation semantics \citep{zhu2021imitation,zhang2023integration}. Consequently, state-of-the-art claims are difficult to audit across papers, even when datasets or model families appear similar. 

In addition, existing comparisons typically ask which model obtains the best aggregate score on a fixed dataset. They less often ask which inductive biases fail under which data-generating regimes. This distinction matters for STPPs: temporal--spatial factorization, self-excitation, smooth latent dynamics, flow-based spatial densities, and diffusion-style generation encode different assumptions about event structure. Real datasets alone rarely isolate these assumptions. \rev{We therefore include a controlled synthetic suite, HawkesNest~\rev{\citep{aalaila2026hawkesnest}}, whose generator exposes several structural axes; here we vary one of them, spatiotemporal entanglement, while holding the background and branching fixed.} This allows us to evaluate whether models degrade gracefully under controlled changes to the data-generating process, rather than only asking which model performs well on average.

We introduce \textsc{Seahorse}, a unified framework for \rev{reproducible training and evaluation of STPPs}. \textsc{Seahorse} standardizes dataset handling, model construction, training modes, hyperparameter tuning, and evaluation artifacts under a single executable benchmarking contract. Models are expressed through a common conceptual pipeline---\rev{a history encoder, a state evolution mechanism, and a decoder that defines the law of the next event}---while still allowing intensity-based, density-based, and generative parameterizations. This enables classical, neural, and diffusion-style STPP models to be evaluated under shared preprocessing, fixed splits, coordinate-aware likelihood reporting, and consistent post-NLL diagnostics. Figure~\ref{fig:seahorse_overview} \rev{illustrates this role: models that answer the same question, when and where the next event occurs, with an event rate, a probability density, or samples are trained and evaluated under one protocol and compared on a common scale.} \rev{The implementation}, including preset configurations and reproduction scripts, is available at \rev{\repo}.
Our contributions are:

\begin{enumerate}
    \item \textbf{A unified execution contract for STPPs.}
    \textsc{Seahorse} provides a modular interface for STPP models based on \rev{a history encoder, a state evolution mechanism, and a decoder}. The same execution layer supports intensity-based, density-based, and generative models, allowing heterogeneous STPP families to share dataset handling, training modes, tuning, and evaluation artifacts.

    \item \textbf{Reproducible cross-family benchmarking.}
    We implement a standardized benchmark protocol covering classical baselines, neural intensity models, density-based models, and generative STPPs\rev{, and apply it to 19 presets on 13 real datasets}. All included models are trained, tuned, and evaluated under common splits, preprocessing, and coordinate-aware reporting, enabling directly auditable comparisons across \rev{the evaluated} model families.

    \item \textbf{Controlled diagnostics.}
    Using real datasets and HawkesNest~\rev{\citep{aalaila2026hawkesnest}}, a controlled synthetic STPP suite, we evaluate not only likelihood fit but also \rev{calibration, sample-based next-event prediction, and ground-truth intensity recovery. These diagnostics show where likelihood rankings and predictive quality disagree, and they allow models without a tractable likelihood to be compared.}

    \item \textbf{An extensible research platform.}
    \textsc{Seahorse} is designed as a reusable platform: practitioners can apply included models to new datasets through configuration, while researchers can add new STPP methods by implementing the common model interface or wrapping existing PyTorch modules. This reduces the reimplementation burden and makes future STPP methods comparable to a shared baseline suite.
\end{enumerate}

\section{Background}\label{sec:background}
\rev{Event data consist of discrete events, each with a time and a location, that arrive irregularly and often cluster. We write an observed sequence up to a} \emph{time horizon} $T \in \mathbb{R}_{\geq 0}$ \rev{as $(t_1, \mathbf{s}_1), \dots, (t_n, \mathbf{s}_n)$,} where $t_i \in [0,T]$ denotes the time of event $i$ and \rev{$\mathbf{s}_i \in \mathbb{R}^2$} its spatial location. \rev{Spatiotemporal point processes (STPPs) provide a principled probabilistic model for such data. An STPP} is a stochastic process that induces a distribution over sequences of events occurring in continuous space and time~\citep{moller2003statistical,rasmussen2018lecture}\rev{; an observed sequence is one} realization (sample) of \rev{it}.

\paragraph{\rev{Next-event law.}}
\rev{The joint density of an event sequence factorizes into the conditional distributions of each event given the events before it, together with the probability that no further event occurs before $T$.} This reduces modeling the joint density over entire sequences to specifying the conditional distribution of the next event given the history \rev{$\mathcal{H}_t = \{(t_i, \mathbf{s}_i) \mid t_i < t\}$, and allows sequences to be sampled} one event after another. The next-event distribution can be specified either directly via a conditional density, \rev{$p(t, \mathbf{s} \mid \mathcal{H}_t)$}, or via an intensity function, \rev{$\lambda(t, \mathbf{s} \mid \mathcal{H}_t)$}, which defines the infinitesimal event rate. Under absolute continuity assumption, both formulations are equivalent \cite{rasmussen2018lecture}. \rev{Many models factorize the density into a temporal part and a spatial part given the time, and some provide only a sampler for the next event.}

\paragraph{\rev{Likelihood.}}
\rev{STPPs are usually fitted and compared through the log-likelihood of an observed sequence, the log-density that the model assigns to it. Written with the intensity, it is}
\[
\rev{\mathcal{L} =
\sum_{i=1}^n \log \lambda(t_i, \mathbf{s}_i \mid \mathcal{H}_{t_i})
- \int_0^T \int_{\mathbb{R}^2} \lambda(t, \mathbf{s} \mid \mathcal{H}_t)\, d\mathbf{s}\,dt,}
\]
and is typically optimized \rev{directly or through} a surrogate objective. \rev{The double integral, the \emph{compensator}, has a closed form only for simple intensities and is otherwise computed numerically, at a cost in time and accuracy. Sampler-only models have no tractable likelihood. Since a density depends on the units of time and space, likelihoods are comparable only in common coordinates (Section~\ref{sec:framework}).}

There are two main approaches to parameterizing \rev{STPPs}:
\paragraph{Classical parametric functions.}
These specify explicit parametric forms for the intensity. For a homogeneous Poisson process, \rev{$\lambda_{\boldsymbol{\theta}}(t,\mathbf{s} \mid \mathcal{H}_t)=\mu$}, where the trainable parameter is $\boldsymbol{\theta}=\{\mu\}$, with $\mu$ the constant background event rate.
For a self-exciting Hawkes process,
\[
\rev{\lambda_{\boldsymbol{\theta}}(t,\mathbf{s} \mid \mathcal{H}_t)
=
\mu+
\sum_{t_i<t}\alpha e^{-\beta(t-t_i)}k_\sigma(\mathbf{s}-\mathbf{s}_i),}
\]
where the trainable parameters are $\boldsymbol{\theta}=\{\mu,\alpha,\beta,\sigma\}$: $\mu$ is the background rate, $\alpha$ the excitation strength, $\beta$ the temporal decay, and $\sigma$ the spatial interaction scale. Here, $k_\sigma$ is a spatial kernel describing how influence decays with distance, for instance a Gaussian kernel.
\paragraph{Neural STPPs.}
Modern approaches replace fixed parametric forms with learned components~\citep{mukherjee2025neural}. Most methods follow a common autoregressive pattern: a \emph{history encoder} maps past events into a latent state $h_i = f_{\mathrm{enc}}(\mathcal{H}_{t_i})$, a \emph{state evolution} propagates it forward $h(t) = f_{\mathrm{evo}}(h_i, t_i, t)$, and a \rev{\emph{decoder}} produces the next-event distribution
\[
\rev{(t_{i+1}, \mathbf{s}_{i+1}) \sim p_\theta(\cdot \mid h(t)).}
\]
Architectures differ in each component: the history encoder may be an RNN~\citep{du2016recurrent}, self-attention~\citep{zhang2020self,zuo2020transformer}, or a set encoder; state evolution may be piecewise constant or governed by a neural ODE~\citep{chen2021neural}; and the \rev{decoder} may factorize into temporal and spatial parts~\rev{\citep{chen2021neural}}, model the joint intensity directly~\rev{\citep{zhou2024automatic}}, or use score-based generative models~\citep{yuan2023spatio}.

\begin{figure}[t]
\centering
\begin{tikzpicture}[
    node distance=0.6cm,
    box/.style={
        rectangle, rounded corners=3pt, draw=black!70, fill=black!5,
        minimum height=1.15cm, minimum width=2.3cm,
        text centered, font=\small\sffamily, text width=2.2cm,
        line width=0.5pt
    },
    io/.style={
        font=\small
    },
    arr/.style={
        -{Stealth[length=5pt]}, thick, black!60
    },
    note/.style={
        font=\scriptsize, text=black!50, text width=2.4cm, text centered
    }
]

\node[io] (input) {Event history};

\node[box, right=of input] (enc) {History\\Encoder};
\node[box, right=of enc]   (evo) {State\\Evolution};
\node[box, right=of evo]   (dec) {Decoder};

\node[io, right=0.5cm of dec] (output) {Next event};

\draw[arr] (input) -- (enc);
\draw[arr] (enc) -- (evo);
\draw[arr] (evo) -- (dec);
\draw[arr] (dec) -- (output);

\node[note, below=0.3cm of enc]
    {RNN, Transformer,\\set encoder};
\node[note, below=0.3cm of evo]
    {Piecewise constant,\\neural ODE + jumps};
\node[note, below=0.3cm of dec]
    {Mixture, CNF,\\integral net, score net};

\end{tikzpicture}
\caption{Unified model decomposition for neural STPPs. Every method in our benchmark instantiates these three components; families differ in the architectural choice at each stage (annotated below). Table~\ref{tab:unified-design-space} gives the per-method assignments.}
\label{fig:pipeline}
\end{figure}

\section{A Unified Framework for Neural STPPs}
\label{sec:design-space}
Figure~\ref{fig:pipeline} summarizes the encode--evolve--decode abstraction used by \textsc{Seahorse}. Given the observed past, each method builds a conditioning representation and maps it to a predictive law for the next time--location pair. The abstraction separates three roles: the history encoder determines what information is retained from the past, the state evolution mechanism determines how this representation changes between events, and the decoder exposes the predictive object used for training and evaluation. This gives intensity-based, density-based, and generative STPPs a common forward contract without forcing them to share the same internal parameterization.
\paragraph{History encoder.}
The history encoder determines what information from the observed past \(\mathcal{H}_{t}\) is available to the model. In classical processes this role is played by hand-designed sufficient statistics, such as cumulative counts, elapsed times, or running excitation terms. Neural models replace these summaries with learned representations: recurrent encoders process events sequentially \citep{du2016recurrent,mei2017neural}, attention-based encoders condition on a finite or full event window \citep{zhang2020self,zuo2020transformer}, and window-based architectures retain per-event embeddings together with raw time--space coordinates. The output may be a single latent state, a set of event embeddings, or conditioning tokens; in all cases, it defines the information about the past that downstream components may use.
\paragraph{State evolution.}
The state evolution component specifies how the conditioning representation changes between observed events. Some models are static with respect to event time: the encoder produces a representation at the most recent event, and this representation is reused until the next event is observed. This covers many window-based likelihood models and score-based next-event generators, whose denoising dynamics are internal to decoding rather than physical-time evolution. Other models use continuous-time state dynamics, as in Neural STPP~\rev{\citep{chen2021neural}}, where a latent state evolves according to an ODE between events and is updated at event times. Continuous evolution can represent smoothly changing inter-event dynamics, but introduces solver cost and numerical tolerances into training and evaluation. In \textsc{Seahorse}, both cases satisfy the same contract: given a history and query time, they provide the conditioning state required by the decoder.
\paragraph{Decoder.}
The decoder maps the current latent state to a law for the next event's time and location, and is where STPP families differ most. {Factorized decoders} combine a temporal model with a conditional spatial model, such as \rev{a recurrent or self-attentive temporal intensity~\citep{du2016recurrent,zuo2020transformer} paired with a Gaussian mixture over locations}, or a hazard-based temporal intensity with a conditional normalizing flow for space~\citep{chen2021neural}. {Joint decoders} model time and space as one object: AutoSTPP~\rev{\citep{zhou2024automatic}} parameterizes a joint intensity through a monotone integral network with an exact compensator, while NSMPP~\rev{\citep{zhu2022neural}} uses spectral features for a joint conditional intensity. {Generative decoders} produce samples rather than explicit densities: SMASH~\citep{li2024beyond} uses a learned score function and Langevin dynamics, while Diffusion STPP~\citep{yuan2023spatio} uses a multi-step diffusion process. The decoder output determines both the training objective and the available evaluation queries: intensity and density decoders support likelihood-based training and pointwise likelihood or intensity evaluation, whereas score-based decoders require surrogate losses and typically provide samples from which likelihood-related quantities must be approximated.

\noindent Table~\ref{tab:unified-design-space} instantiates this decomposition for \rev{the neural methods in our benchmark; the classical baselines are summarized in Appendix~\ref{app:benchmark_model_summaries}}. The unification is not a shared formula but a stable interface: each component has a well-defined input/output contract, and families differ only in how they fill each slot. This is what enables \textsc{Seahorse} to train, evaluate, and compare all methods through a single pipeline.

\section{Seahorse: A Framework for Heterogeneous STPPs}
\label{sec:framework}
The unified abstraction in Section~\ref{sec:design-space} is not sufficient by itself to make STPP results comparable. Two models may both report NLL while evaluating different quantities: one may score events in normalized coordinates, another in raw coordinates; one may optimize an exact point-process likelihood, another a variational or score-matching surrogate. \textsc{Seahorse} addresses this by separating a model's internal training space from the benchmark reporting space\rev{: models} may transform and parameterize the data internally, but benchmark metrics are reported under a shared raw-space protocol with the required coordinate correction. The \rev{repository at \repo{}} provides the corresponding implementation.
\begin{table*}[t]
\centering
\footnotesize
\setlength{\tabcolsep}{4pt}
\caption{
Neural and neural-style methods covered by the unified STPP interface.
Classical factorized and temporal-only baselines are summarized in Appendix~\ref{app:benchmark_model_summaries}.
}
\label{tab:unified-design-space}
\begin{tabularx}{\textwidth}{@{}l Y Y Y l@{}}
\toprule
\textbf{Method} & \textbf{\rev{History encoder}} & \textbf{\rev{State evolution}} & \textbf{\rev{Decoder}} & \textbf{Objective} \\
\midrule
DeepSTPP~\citep{zhou2022neural} & \leavevmode\rev{Variational encoder} & Latent stochastic process & \leavevmode\rev{Nonparametric joint intensity} & \leavevmode\rev{Variational bound} \\
AutoSTPP~\citep{zhou2024automatic} & \leavevmode\rev{Fixed event window} & \leavevmode\rev{Integral network} & \leavevmode\rev{Joint intensity, exact integral} & \leavevmode\rev{Exact likelihood} \\
NSMPP~\rev{\citep{zhu2022neural}} & Raw history & \leavevmode\rev{None} & \leavevmode\rev{Joint intensity, quadrature} & \leavevmode\rev{Likelihood} \\
NJSDE~\citep{jia2019neural} & \leavevmode\rev{Recurrent} & \leavevmode\rev{Neural ODE} & \leavevmode\rev{Temporal intensity + GMM} & \leavevmode\rev{Likelihood + reg.} \\
Jump-CNF~\citep{chen2021neural} & \leavevmode\rev{Recurrent} & \leavevmode\rev{Neural ODE with jumps} & \leavevmode\rev{Temporal intensity + jump CNF} & \leavevmode\rev{Likelihood + reg.} \\
Attn-CNF~\citep{chen2021neural} & \leavevmode\rev{Recurrent / attention} & \leavevmode\rev{Neural ODE, path-conditioned} & \leavevmode\rev{Temporal intensity + attentive CNF} & \leavevmode\rev{Likelihood + reg.} \\
SMASH~\citep{li2024beyond} & \leavevmode\rev{Conditional score network} & \leavevmode\rev{Static (dynamics in decoder)} & \leavevmode\rev{Score-based sampler} & \leavevmode\rev{Score matching} \\
DSTPP~\citep{yuan2023spatio} & \leavevmode\rev{Spatio-temporal co-attention} & \leavevmode\rev{Static (dynamics in decoder)} & \leavevmode\rev{Diffusion sampler} & \leavevmode\rev{Denoising (ELBO)} \\
\bottomrule
\end{tabularx}
\end{table*}

\begin{figure}
    \centering
    \includegraphics[width=\linewidth]{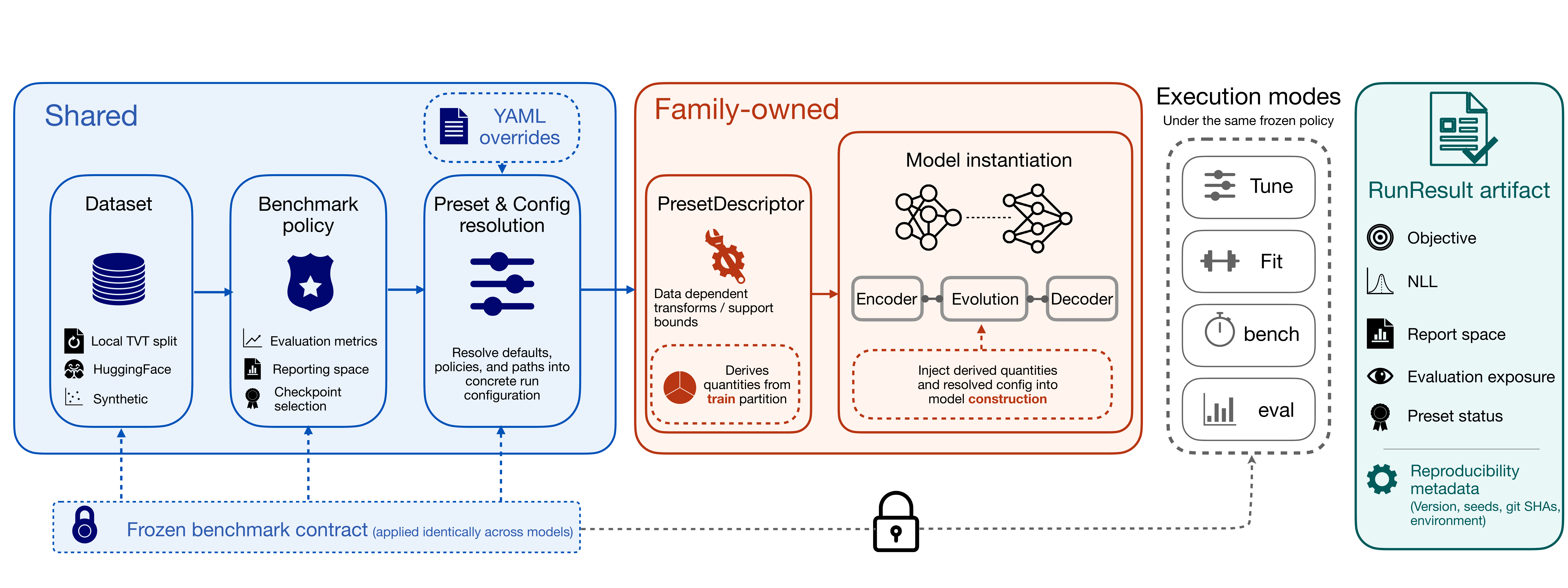}
\caption{
\textbf{\textsc{Seahorse} execution contract.}
Shared benchmark components fix the dataset, split, policy, and resolved configuration, while family-owned presets inject training-only support statistics and transforms before model construction.
All execution modes preserve the same policy metadata and write a \textsc{RunResult} artifact for auditable reporting.
}
    \label{fig:seahorse-contract}
\end{figure}

\paragraph{Benchmark contract.}
\textsc{Seahorse} uses a raw-first benchmark contract. Datasets, event order, and train/validation/test splits are fixed in original data coordinates. A model preset may estimate its own normalization, coordinate transform, support bounds, or quadrature domain, but only from the training partition, and the resulting transform is stored with the run. Evaluation is fixed separately from training: test NLL is computed event-wise by conditioning on the realized pre-event history and scoring the next event in raw coordinates. \rev{All models are scored on the same target events, each conditioned on the same number of preceding events (Appendix~\ref{sec:table-notes}).} \rev{If a model evaluates its likelihood in transformed coordinates $u=g(z)$ of the raw event $z=(\Delta t,\mathbf{s})$, \textsc{Seahorse} reports $\mathrm{NLL}_{\mathrm{raw}}=\mathrm{NLL}_{\mathrm{native}}-\log\lvert\det \partial g/\partial z\rvert$ (Appendix, Eq.~\ref{eq:jacobian-correction}).} Thus, validation objectives may differ across families, but reported benchmark NLLs remain tied to the same event-level target and reference measure.

\paragraph{Model instantiation.}
As shown in Figure~\ref{fig:seahorse-contract}, benchmark runs are specified by named configurations identifying the dataset, model family, training policy, and hyperparameters. Instantiation follows a fixed order: \textsc{Seahorse} resolves the benchmark policy, applies configuration overrides, calls the family-owned preset to derive training-data-dependent quantities, and constructs the executable model \rev{from} the components defined by the unified interface in Section~\ref{sec:design-space}. This prevents support bounds, coordinate transforms, and normalization statistics from leaking information from validation or test data.

\paragraph{Training, sampling, and tuning.}
Training remains family-specific\rev{: each} preset optimizes the objective for which the method was designed, including exact likelihoods, numerical likelihoods, variational objectives, score-matching losses, or auxiliary regularized objectives. Sampling is also capability-dependent: generative models use their native samplers, while intensity-based models can sample through thinning or model-specific next-event routines where available. Hyperparameter search is exposed through the \texttt{tune} mode, with Ray Tune support for benchmark-scale sweeps. \rev{The} training objective, sampling path, and evaluation exposure \rev{are recorded} in the run metadata.

\paragraph{Execution pipeline.}
\textsc{Seahorse} exposes four \rev{core} execution modes: \texttt{fit} trains one configured model; \texttt{tune} performs hyperparameter search; \texttt{bench} launches multi-model, multi-dataset, multi-seed benchmark campaigns under a frozen policy; and \texttt{evaluate} computes post-fit diagnostics. Each run artifact records the training objective, likelihood reporting convention, reporting coordinate space, supported evaluation modalities, selected checkpoint, and preset status. This makes downstream tables auditable: likelihoods, predictive scores, and diagnostics remain tied to the objective, coordinate space, and evaluation interface that produced them.

\paragraph{\rev{Software interface.}}\label{sec:software-interface}
\textsc{Seahorse} exposes a command-line interface for large-scale execution and a Python interface for interactive use, both backed by the same configuration and runtime contracts. Practitioners can run included models on benchmark datasets or on new datasets following the \textsc{Seahorse} data schema. Researchers can add new models by registering a preset or wrapping an existing PyTorch module, after which the model can be trained, tuned, and evaluated against the same datasets and metrics as the included baselines. \rev{The package is open source under the Apache-2.0 license (\texttt{pip install seahorse-stpp}), with 28 model presets, including the 19 benchmarked in this paper, more than 500 automated tests run in continuous integration, and loaders for all datasets used here through the Hugging Face Hub with pinned revisions and hash-verified splits.} Full configuration details, API references, dataset specifications, and extension examples are provided in the documentation \rev{at \docs}.

\section{Experiments}
\label{sec:experiments}

\subsection{Experimental Setup}
\label{sec:experimental-setup}
\paragraph{Protocol.}
All models are trained and evaluated on identical raw event sequences with fixed train/validation/test splits, averaged over three seeds. We use two evaluation scopes. On real datasets, we run the full benchmark suite and report test NLL across all model families\rev{, together with calibration and next-event diagnostics}. On \rev{the HawkesNest entanglement suite, we compare eight presets against the known data-generating process}. \rev{On real data, every gradient-trained preset is given the same budget: three hyperparameter trials of 5{,}000 optimization steps selected on validation NLL, followed by refits of at most 20{,}000 steps with early stopping; runs cut short by the 72-hour wall-clock limit are marked $\dag$ in Table~\ref{tab:real_data_expanded_nll}.}

\paragraph{Baselines.}
We evaluate the model families summarized in Table~\ref{tab:unified-design-space}\rev{, together with classical and factorized baselines, 19 presets in total (Table~\ref{tab:model-aliases})}. The synthetic comparison focuses on \rev{AutoSTPP, DeepSTPP, NSMPP, NJSDE, SMASH, and DSTPP, with RMTPP and THP as factorized baselines; Attn-CNF and Jump-CNF are omitted there because of their training cost}. Implementation details are in Appendix~\ref{app:benchmark_model_summaries}.

\paragraph{Datasets.}
Real datasets \rev{comprise ten event datasets with chronological splits, namely earthquake catalogs (ComCat, GeoNet), bike-share trips (Citi Bike, hourly sequences), 311 calls (Austin, NYC), 911 calls (Seattle), traffic collisions (NYC), crime reports (Chicago, LA), and ride-hailing pickups (Uber), and the three datasets of the Neural STPP benchmark~\citep{chen2021neural} (Citi Bike with daily sequences, Earthquakes, and COVID) under their original splits}. Synthetic data is generated with HawkesNest\rev{~\citep{aalaila2026hawkesnest}}, our controlled STPP generator, \rev{here varying spatiotemporal entanglement over four levels while holding the background and branching fixed, and storing the ground-truth intensity} for direct recovery diagnostics. Preprocessing details are in Appendix~\ref{app:datasets}.

\paragraph{Evaluation.}
The primary metric is held-out per-event test NLL in raw coordinates; for models trained in normalized coordinates we apply the corresponding log-Jacobian correction so that likelihoods are directly comparable. Beyond NLL, we report \rev{calibration of the predicted waiting times and locations, sample-based next-event scores, and, on the synthetic suite, excess NLL over the true process and ground-truth intensity recovery}. Diagnostic definitions are in Appendix~\ref{app:evaluation-semantics}.

\subsection{Results and discussion}

\begin{figure}[t]
\centering
\includegraphics[width=\linewidth]{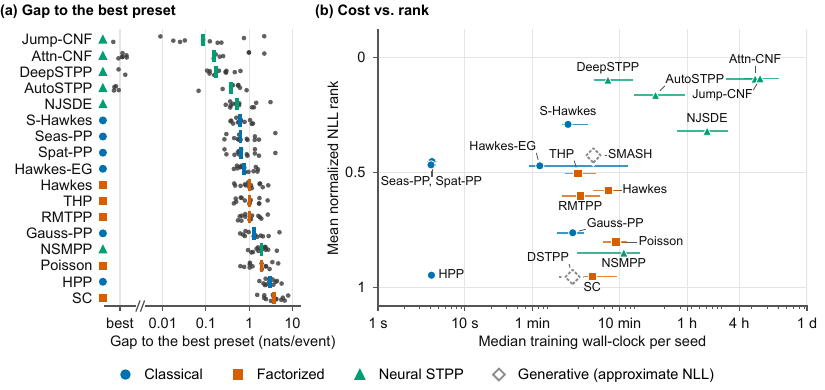}
\caption{\rev{\textbf{Real-data results.} (a) Gap in test NLL to the best preset on each of the 13 real datasets (nats/event, log scale); dot: one dataset, bar: median over datasets, ``best'': zero gap. (b) Median training time per seed (wall-clock on heterogeneous GPUs; bar: interquartile range) against the mean normalized within-dataset NLL rank (0 = best). The generative models (hollow markers) are placed by approximate likelihoods, an upper bound for DSTPP and a subsample estimate for SMASH, for orientation only.}}
\label{fig:main_results}
\end{figure}

\paragraph{Real-data likelihood.}
\rev{Table~\ref{tab:real_data_expanded_nll} reports test NLL on the thirteen real datasets, and Figure~\ref{fig:main_results} summarizes the gaps to the best model and the training cost (within-dataset ranks: Appendix, Figure~\ref{fig:rank_overview}).} The ranking is domain-dependent: Attn-CNF is strongest on \rev{COVID, Citi Bike (d) and four of the urban datasets, DeepSTPP on Earthquakes and Austin, and AutoSTPP on ComCat, GeoNet, Citi Bike (h) and NYC~311, while Jump-CNF is best on Uber.} \rev{The classical self-exciting S-Hawkes outperforms NSMPP on every dataset}, indicating that classical self-excitation \rev{can still match or exceed neural methods in some regimes}. Poisson and self-correcting variants are consistently weaker. \rev{Cost changes the picture (Figure~\ref{fig:main_results}b): DeepSTPP is within a median 0.17 nats of the best model at about seven minutes of training per seed, while Attn-CNF and Jump-CNF reach 0.15 and 0.09 at six to seven hours, and Jump-CNF diverged on two datasets.}

\begin{table*}[t]
\centering
\caption{Test next-event NLL (nats per event, lower is better), mean $\pm$ std over three seeds; values are comparable only within a column. Bold and underline: best and second-best value per dataset. \emph{Generative} models have no exact likelihood; their values (grey) are a variational bound (DSTPP) or an approximation (SMASH) and are not ranked. $\dag$: reported with a caveat (fewer seeds, budget limit, or a failed numerical check); per-cell details in Table~\ref{tab:real_data_full_nll}.}
\label{tab:real_data_expanded_nll}
\scriptsize
\setlength{\tabcolsep}{2pt}
\begin{tabular}{lccccccc}
\toprule
\textbf{Preset} & \textbf{COVID} & \textbf{Earthquakes} & \textbf{Citi Bike (d)} & \textbf{ComCat} & \textbf{GeoNet} & \textbf{Citi Bike (h)} & \textbf{Austin} \\
\midrule
\multicolumn{8}{l}{\emph{Classical}} \\
HPP & $-0.193_{\pm0.00}$ & $6.405_{\pm0.00}$ & $-5.257_{\pm0.00}$ & $5.155_{\pm0.00}$ & $3.823_{\pm0.00}$ & $-11.326_{\pm0.00}$ & $-6.537_{\pm0.00}$ \\
Seas-PP & $-2.063_{\pm0.00}$ & $5.096_{\pm0.00}$ & $-8.752_{\pm0.00}$ & $3.576_{\pm0.00}$ & $0.863_{\pm0.00}$ & $-14.804_{\pm0.00}$ & $-8.307_{\pm0.00}$ \\
Spat-PP & $-2.056_{\pm0.00}$ & $5.093_{\pm0.00}$ & $-8.752_{\pm0.00}$ & $3.573_{\pm0.00}$ & $0.862_{\pm0.00}$ & $-14.805_{\pm0.00}$ & $-8.290_{\pm0.00}$ \\
Hawkes-EG & $-2.070_{\pm0.00}$ & $4.378_{\pm0.00}$ & $-8.813_{\pm0.00}$ & $0.796_{\pm0.00}$ & $0.980_{\pm0.00}$ & $-14.946_{\pm0.00}$ & $-8.400_{\pm0.00}$ \\
Gauss-PP & $-1.238_{\pm0.04}$ & $6.231_{\pm0.00}$ & $-8.718_{\pm0.00}$ & $4.557_{\pm0.01}$ & $1.849_{\pm0.01}$ & $-14.740_{\pm0.00}$ & $-8.118_{\pm0.01}$ \\
S-Hawkes & $-2.143_{\pm0.02}$ & $4.558_{\pm0.01}$ & $-8.910_{\pm0.00}$ & $0.579_{\pm0.00}$ & $0.940_{\pm0.00}$ & $-15.073_{\pm0.00}$ & $-8.556_{\pm0.00}$ \\
\midrule
\multicolumn{8}{l}{\emph{Factorized}} \\
Poisson & $-1.471_{\pm0.01}$ & $5.374_{\pm0.00}$ & $-8.707_{\pm0.00}$ & $2.979_{\pm0.00}$ & $1.342_{\pm0.00}$ & $-12.753_{\pm0.00}$ & $-7.744_{\pm0.00}$ \\
Hawkes & $-1.556_{\pm0.01}$ & $5.188_{\pm0.00}$ & $-8.861_{\pm0.00}$ & $2.302_{\pm0.01}$ & $1.327_{\pm0.00}$ & $-14.883_{\pm0.00}$ & $-8.490_{\pm0.00}$ \\
SC & $3.020_{\pm0.05}$ & $6.614_{\pm0.01}$ & $-7.094_{\pm0.02}$ & $3.083_{\pm0.00}$ & $1.848_{\pm0.01}$ & $-9.061_{\pm0.00}$ & $-4.669_{\pm0.01}$ \\
RMTPP & $-1.841_{\pm0.03}$ & $5.331_{\pm0.00}$ & $-8.785_{\pm0.00}$ & $2.229_{\pm0.01}$ & $1.335_{\pm0.00}$ & $-14.903_{\pm0.00}$ & $-8.336_{\pm0.00}$ \\
THP & $-2.012_{\pm0.01}$ & $5.117_{\pm0.00}$ & $-8.846_{\pm0.00}$ & $2.206_{\pm0.01}$ & $1.331_{\pm0.00}$ & $-15.048_{\pm0.00}$ & $-8.498_{\pm0.00}$ \\
\midrule
\multicolumn{8}{l}{\emph{Neural STPP}} \\
AutoSTPP & $-2.128_{\pm0.02}$ & \underline{$3.655_{\pm0.01}$} & $-8.913_{\pm0.02}$ & $\mathbf{-0.469_{\pm0.06}}$ & $\mathbf{0.315_{\pm0.14}}$ & $\mathbf{-15.717_{\pm0.20}}$ & \underline{$-8.873_{\pm0.03}$} \\
DeepSTPP & $-2.391_{\pm0.08}$ & $\mathbf{3.411_{\pm0.03}}$ & $-9.214_{\pm0.02}$ & \underline{$-0.004_{\pm0.06}$} & $0.421_{\pm0.02}$ & $-15.433_{\pm0.01}$ & $\mathbf{-8.941_{\pm0.01}}$ \\
NSMPP & $-1.291_{\pm0.97}^{\dag}$ & $5.775_{\pm0.71}$ & $-7.014_{\pm1.52}^{\dag}$ & $3.898_{\pm1.08}$ & $1.954_{\pm1.62}$ & $-13.339_{\pm1.85}$ & $-7.454_{\pm0.78}$ \\
NJSDE & $-2.226_{\pm0.02}$ & $4.865_{\pm0.51}$ & $-8.960_{\pm0.01}$ & $2.646_{\pm0.26}$ & $1.473_{\pm0.02}$ & $-15.126_{\pm0.02}$ & $-8.577_{\pm0.01}$ \\
Attn-CNF & $\mathbf{-2.513_{\pm0.05}}$ & $4.397_{\pm0.09}$ & $\mathbf{-9.335_{\pm0.01}}$ & $1.773_{\pm0.04}^{\dag}$ & $0.466_{\pm0.00}^{\dag}$ & \underline{$-15.467_{\pm0.02}$} & $-8.773_{\pm0.02}$ \\
Jump-CNF & \underline{$-2.496_{\pm0.01}$} & $4.169_{\pm0.07}^{\dag}$ & \underline{$-9.326_{\pm0.01}^{\dag}$} & $1.762_{\pm0.05}^{\dag}$ & \underline{$0.340_{\pm0.03}$} & {\scriptsize diverged} & $-8.712_{\pm0.01}$ \\
\midrule
\multicolumn{8}{l}{\emph{Generative}} \\
SMASH & \cellcolor{gray!12}$-2.365_{\pm0.08}$ & \cellcolor{gray!12}$6.293_{\pm0.59}$ & \cellcolor{gray!12}$-8.942_{\pm0.05}$ & \cellcolor{gray!12}$2.193_{\pm0.35}$ & \cellcolor{gray!12}$1.725_{\pm0.16}$ & \cellcolor{gray!12}$-15.177_{\pm0.02}$ & \cellcolor{gray!12}$-8.722_{\pm0.05}$ \\
DSTPP & \cellcolor{gray!12}$0.874_{\pm0.54}$ & \cellcolor{gray!12}$7.970_{\pm0.48}$ & \cellcolor{gray!12}$-6.624_{\pm0.16}$ & \cellcolor{gray!12}$148.691_{\pm0.09}^{\dag}$ & \cellcolor{gray!12}$3.414_{\pm0.40}$ & \cellcolor{gray!12}$-10.739_{\pm0.17}$ & \cellcolor{gray!12}$-5.776_{\pm0.25}$ \\
\bottomrule
\end{tabular}

\vspace{2pt}
\begin{tabular}{lcccccc}
\toprule
\textbf{Preset} & \textbf{Seattle} & \textbf{NYC 311} & \textbf{NYC coll.} & \textbf{Chicago} & \textbf{LA} & \textbf{Uber} \\
\midrule
\multicolumn{7}{l}{\emph{Classical}} \\
HPP & $-7.100_{\pm0.00}$ & $-9.227_{\pm0.00}$ & $-5.627_{\pm0.00}$ & $-6.764_{\pm0.00}$ & $-6.241_{\pm0.00}$ & $-10.476_{\pm0.00}$ \\
Seas-PP & $-8.944_{\pm0.00}$ & $-10.854_{\pm0.00}$ & $-7.081_{\pm0.00}$ & $-8.384_{\pm0.00}$ & $-7.806_{\pm0.00}$ & $-13.529_{\pm0.00}$ \\
Spat-PP & $-8.943_{\pm0.00}$ & $-10.850_{\pm0.00}$ & $-7.081_{\pm0.00}$ & $-8.384_{\pm0.00}$ & $-7.805_{\pm0.00}$ & $-13.530_{\pm0.00}$ \\
Hawkes-EG & $-8.439_{\pm0.00}$ & $-10.609_{\pm0.00}$ & $-6.710_{\pm0.00}$ & $-7.839_{\pm0.00}$ & $-7.608_{\pm0.00}$ & $-13.484_{\pm0.00}$ \\
Gauss-PP & $-8.444_{\pm0.00}$ & $-10.353_{\pm0.00}$ & $-6.629_{\pm0.00}$ & $-7.981_{\pm0.00}$ & $-7.081_{\pm0.00}$ & $-12.724_{\pm0.01}$ \\
S-Hawkes & $-8.691_{\pm0.00}$ & $-11.569_{\pm0.04}$ & $-6.767_{\pm0.00}$ & $-8.480_{\pm0.01}$ & $-7.938_{\pm0.00}$ & $-13.610_{\pm0.00}$ \\
\midrule
\multicolumn{7}{l}{\emph{Factorized}} \\
Poisson & $-8.209_{\pm0.00}$ & $-7.502_{\pm0.00}$ & $-6.583_{\pm0.00}$ & $-7.327_{\pm0.00}$ & $-6.744_{\pm0.00}$ & $-11.649_{\pm0.00}$ \\
Hawkes & $-8.443_{\pm0.00}$ & $-10.143_{\pm0.00}$ & $-6.717_{\pm0.00}$ & $-8.114_{\pm0.00}$ & $-7.414_{\pm0.00}$ & $-13.008_{\pm0.01}$ \\
SC & $-7.018_{\pm0.01}$ & $-4.207_{\pm0.01}$ & $-4.895_{\pm0.00}$ & $-5.579_{\pm0.01}$ & $-4.999_{\pm0.00}$ & $-8.107_{\pm0.00}$ \\
RMTPP & $-8.425_{\pm0.00}$ & $-10.432_{\pm0.00}$ & $-6.686_{\pm0.00}$ & $-8.165_{\pm0.00}$ & $-7.532_{\pm0.00}$ & $-12.943_{\pm0.00}$ \\
THP & $-8.444_{\pm0.00}$ & $-10.504_{\pm0.01}$ & $-6.710_{\pm0.00}$ & $-8.180_{\pm0.00}$ & $-7.433_{\pm0.03}$ & $-13.081_{\pm0.01}$ \\
\midrule
\multicolumn{7}{l}{\emph{Neural STPP}} \\
AutoSTPP & $-8.678_{\pm0.03}$ & $\mathbf{-12.203_{\pm0.48}}$ & $-6.796_{\pm0.01}$ & $-8.649_{\pm0.04}$ & $-8.291_{\pm0.06}$ & $-13.730_{\pm0.01}$ \\
DeepSTPP & $-9.085_{\pm0.04}$ & \underline{$-11.573_{\pm0.01}$} & $-7.177_{\pm0.02}$ & $-9.064_{\pm0.02}$ & $-8.871_{\pm0.08}$ & $-13.816_{\pm0.40}$ \\
NSMPP & $-8.115_{\pm0.88}$ & $-10.266_{\pm0.90}$ & $-6.407_{\pm0.74}$ & $-7.659_{\pm0.81}$ & $-6.409_{\pm0.29}$ & $-11.703_{\pm1.05}$ \\
NJSDE & $-8.656_{\pm0.00}$ & $-10.913_{\pm0.14}$ & $-6.925_{\pm0.00}$ & $-8.900_{\pm0.01}^{\dag}$ & $-8.637_{\pm0.07}^{\dag}$ & $-13.616_{\pm0.00}$ \\
Attn-CNF & $\mathbf{-9.241_{\pm0.00}}$ & $-11.213_{\pm0.01}$ & $\mathbf{-7.349_{\pm0.02}}$ & $\mathbf{-9.264_{\pm0.05}^{\dag}}$ & $\mathbf{-9.163_{\pm0.05}^{\dag}}$ & \underline{$-13.906_{\pm0.30}$} \\
Jump-CNF & \underline{$-9.208_{\pm0.03}^{\dag}$} & {\scriptsize diverged} & \underline{$-7.226_{\pm0.02}^{\dag}$} & \underline{$-9.178_{\pm0.01}^{\dag}$} & \underline{$-8.976_{\pm0.03}$} & $\mathbf{-14.114_{\pm0.05}^{\dag}}$ \\
\midrule
\multicolumn{7}{l}{\emph{Generative}} \\
SMASH & \cellcolor{gray!12}$-8.907_{\pm0.06}$ & \cellcolor{gray!12}$-9.942_{\pm1.14}$ & \cellcolor{gray!12}$-6.522_{\pm0.58}$ & \cellcolor{gray!12}$-8.861_{\pm0.04}$ & \cellcolor{gray!12}$-8.409_{\pm0.13}$ & \cellcolor{gray!12}$-13.596_{\pm0.09}$ \\
DSTPP & \cellcolor{gray!12}$-5.949_{\pm0.28}$ & \cellcolor{gray!12}$-7.625_{\pm0.13}$ & \cellcolor{gray!12}$-4.518_{\pm0.10}$ & \cellcolor{gray!12}$-5.600_{\pm0.29}$ & \cellcolor{gray!12}$-5.136_{\pm0.25}$ & \cellcolor{gray!12}$-9.756_{\pm0.12}$ \\
\bottomrule
\end{tabular}
\end{table*}

\paragraph{\rev{Calibration and next-event prediction.}}
\rev{On the ten benchmark datasets, the best-calibrated model (lowest temporal PIT statistic, Table~\ref{tab:calibration_main}) differs from the likelihood winner on at least seven. Sample-based energy scores of the next event (Table~\ref{tab:nextevent_main}) mostly preserve the likelihood ranking (median Spearman correlation 0.88), with exceptions: DeepSTPP has the best mean rank (2.1 of 19), AutoSTPP, among the top three under NLL, is mid-field (8.3), and SMASH, which has no exact likelihood, ranks ahead of it (7.2).}

\paragraph{Controlled entanglement.}
Table~\ref{tab:suite3_entanglement_test_nll} reports NLL on the HawkesNest entanglement suite, where levels \(\mathbf{L}_0\)--\(\mathbf{L}_3\) progressively strengthen spatiotemporal coupling in the data-generating process. \rev{Because the best achievable NLL changes with the level, we report the excess over the true process; the first row is the excess of the best separable Hawkes fit. AutoSTPP} achieves the best likelihood at every level, consistent with its inductive bias: \rev{it parameterizes the joint intensity over \((t,\mathbf{s})\) directly, with an exact compensator, and so} avoids the temporal--spatial separation imposed by factorized models. \rev{It is the only model whose excess stays below the separable reference at \(\mathbf{L}_1\)--\(\mathbf{L}_3\), indicating that it captures part of the coupling. DeepSTPP is second at every level, but its excess grows from 0.21 to 0.63, faster than the separable reference. NSMPP, NJSDE and the factorized baselines are between 0.9 and 2.4 nats from the true process at every level, and their excess rises by at most 0.15 nats from \(\mathbf{L}_0\) to \(\mathbf{L}_3\). For DSTPP only an upper bound is available, and SMASH has no likelihood.}

\begin{figure*}[t]
\centering
\begin{subfigure}[t]{0.245\textwidth}
    \centering
    \includegraphics[width=\linewidth]{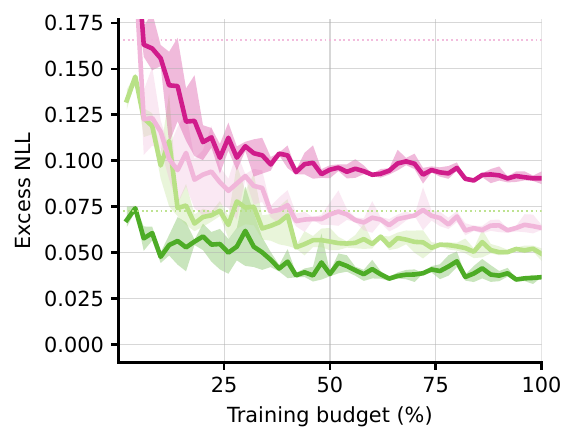}
    \caption{AutoSTPP}
    \label{fig:suite3-bud-autostpp}
\end{subfigure}\hfill
\begin{subfigure}[t]{0.245\textwidth}
    \centering
    \includegraphics[width=\linewidth]{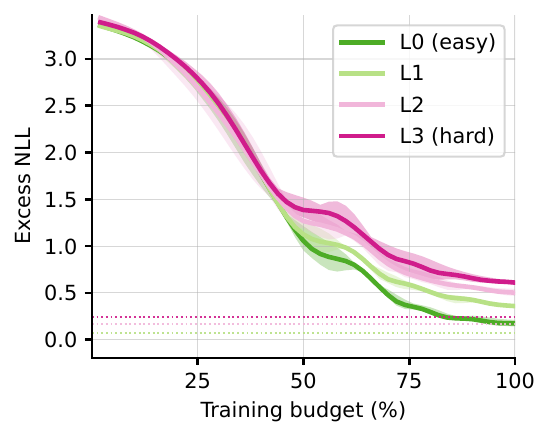}
    \caption{DeepSTPP}
    \label{fig:suite3-bud-deepstpp}
\end{subfigure}\hfill
\begin{subfigure}[t]{0.245\textwidth}
    \centering
    \includegraphics[width=\linewidth]{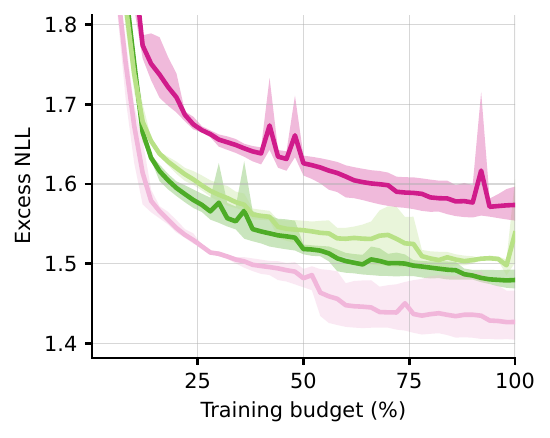}
    \caption{NSMPP}
    \label{fig:suite3-bud-nsmpp}
\end{subfigure}\hfill
\begin{subfigure}[t]{0.245\textwidth}
    \centering
    \includegraphics[width=\linewidth]{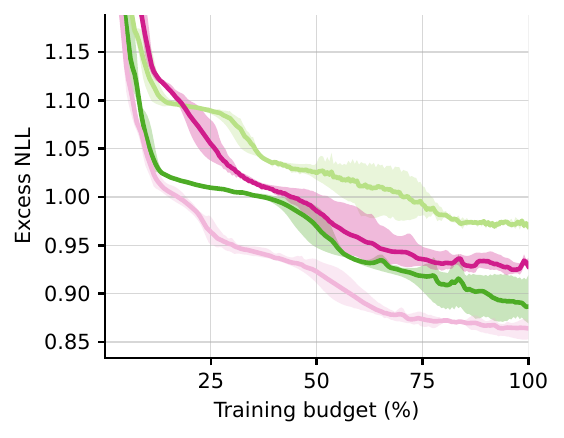}
    \caption{\rev{NJSDE}}
    \label{fig:suite3-bud-njsde}
\end{subfigure}

\caption{
Learning dynamics on the HawkesNest entanglement suite. Each panel reports \rev{validation NLL in excess of the true process} as a function of training budget for one model, with curves corresponding to increasing entanglement levels \(\mathbf{L}_0\)--\(\mathbf{L}_3\)\rev{; dotted lines in (a) and (b) mark the separable reference of each level (off scale where it exceeds the axis range)}.
}
\label{fig:suite3_budget_curves}
\end{figure*}

\begin{table}[t]
\begin{minipage}[t]{0.52\linewidth}\vspace{0pt}
\centering
\captionof{table}{\rev{HawkesNest entanglement suite: test NLL in excess of the true process (nats/event), mean over 3 seeds; last column with 95\% CI. Sep.\ oracle: best separable Hawkes fit. Bold and underline: best and second best. Grey: generative models, not ranked (DSTPP: variational bound).}}
\label{tab:suite3_entanglement_test_nll}
{\footnotesize\setlength{\tabcolsep}{4pt}%
\begin{tabular}{@{}lccccc@{}}
\toprule
Model & L0 & L1 & L2 & L3 & L3$-$L0 \\
\midrule
Sep.\ oracle & $0.00$ & $0.08$ & $0.17$ & $0.25$ & $+0.25_{[.24,.26]}$ \\
\midrule
AutoSTPP & $\mathbf{0.04}$ & $\mathbf{0.06}$ & $\mathbf{0.07}$ & $\mathbf{0.10}$ & $+0.06_{[.05,.08]}$ \\
DeepSTPP & $\underline{0.21}$ & $\underline{0.35}$ & $\underline{0.50}$ & $\underline{0.63}$ & $+0.43_{[.39,.46]}$ \\
NSMPP & $1.45$ & $1.46$ & $1.50$ & $1.56$ & $+0.11_{[.04,.17]}$ \\
NJSDE & $0.96$ & $0.97$ & $0.93$ & $0.95$ & $-0.01_{[-.05,.03]}$ \\
\midrule
RMTPP & $2.25$ & $2.29$ & $2.32$ & $2.40$ & $+0.15_{[.04,.26]}$ \\
THP & $2.14$ & $2.18$ & $2.21$ & $2.29$ & $+0.15_{[.04,.26]}$ \\
\midrule
DSTPP & \cellcolor{gray!12}$3.29$ & \cellcolor{gray!12}$3.32$ & \cellcolor{gray!12}$3.33$ & \cellcolor{gray!12}$3.35$ & \cellcolor{gray!12}$+0.06_{[-.04,.15]}$ \\
SMASH & \multicolumn{5}{c}{\cellcolor{gray!12}no likelihood} \\
\bottomrule
\end{tabular}}

\end{minipage}\hfill
\begin{minipage}[t]{0.45\linewidth}\vspace{0pt}
\centering
\includegraphics[width=\linewidth]{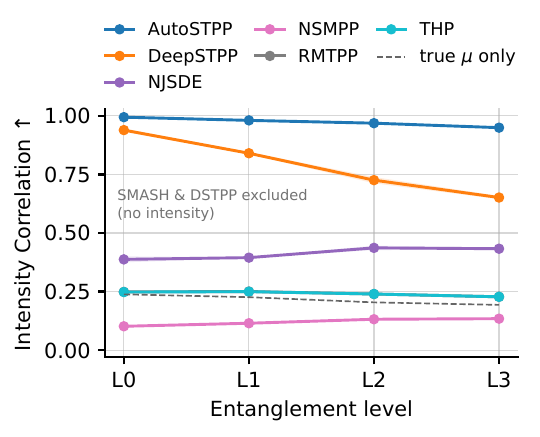}
\captionof{figure}{\rev{Ground-truth intensity recovery on the HawkesNest entanglement suite: correlation between the model and the true intensity across levels (mean over 3 seeds). Dashed: a model that knows only the true background.}}
\label{fig:post-nll-suite3}
\end{minipage}
\end{table}

\paragraph{Learning under increasing entanglement.}
Figure~\ref{fig:suite3_budget_curves} shows \rev{validation NLL in excess of the true process} versus training budget for the same entanglement sweep. The curves expose behavior hidden by final NLL: stronger coupling changes both the final likelihood and the path by which architectures reach it. AutoSTPP in Figure~\ref{fig:suite3-bud-autostpp} maintains a persistent \(\mathbf{L}_3\) gap\rev{, but its curves separate by level early and end below the separable reference of their level, so the remaining gap is the part of the coupling it does not absorb}. DeepSTPP in Figure~\ref{fig:suite3-bud-deepstpp} \rev{separates by level in the second half of training and is still improving when its budget ends, so its final values depend on the budget}. NSMPP in Figure~\ref{fig:suite3-bud-nsmpp} is the contrast case: its curves remain tightly packed\rev{, consistent with an excess that is large at every level} in Table~\ref{tab:suite3_entanglement_test_nll}. \rev{NJSDE in Figure~\ref{fig:suite3-bud-njsde} behaves similarly. SMASH is not shown because its training objective is not a likelihood.}

\paragraph{Beyond likelihood.}
Figure~\ref{fig:post-nll-suite3} evaluates the same entanglement sweep beyond NLL. \rev{Temporal CRPS is flat across levels for every model (Appendix, Table~\ref{tab:suite3_v2_pit}), so one-step timing scores do not register the coupling. Intensity recovery} shows a different pattern: AutoSTPP and DeepSTPP retain the strongest correlations with the ground-truth intensity surface, \rev{but DeepSTPP's correlation falls from 0.94 to 0.65 as the coupling grows while AutoSTPP's stays above 0.95; RMTPP and THP stay at the level of a background-only model, NJSDE somewhat above it and NSMPP below}.

\section{Related Work}\label{sec:RW}

\paragraph{Neural spatiotemporal point processes.}
Neural point process models have increased the flexibility of event-sequence
modeling, first mainly in the temporal setting through recurrent or
attention-based conditional intensities such as Neural Hawkes, SAHP, and THP
\citep{mei2017neural,zhang2020self,zuo2020transformer}. Recent work extends
this direction to spatiotemporal point processes, where each event carries both
time and location \citep{mukherjee2025neural,cheng2025deep}. Representative
models include neural spatiotemporal event dynamics with learned intensity
components \citep{zhou2022neural}, Neural STPP with continuous-time latent
dynamics and conditional normalizing flows \citep{chen2021neural}, automatically
integrable neural STPPs \citep{zhou2024automatic}, spectral marked point-process
models \citep{zhu2022neural}, and diffusion-based spatiotemporal point processes
\citep{yuan2023spatio}. These methods differ in parameterization, training
objective, likelihood estimation, and computational requirements, making direct
comparison difficult without a shared experimental protocol.

\paragraph{Benchmarking and software for point processes.}
Standardized tooling has received more attention for temporal point processes
than for full spatiotemporal models. EasyTPP provides open benchmarking
infrastructure for temporal point processes, including datasets, model
implementations, evaluation programs, documentation, and unified interfaces
\citep{xue2024easytpp}. HoTPP extends temporal point-process benchmarking toward
long-horizon prediction for marked event sequences \citep{karpukhin2024hotpp}.
These benchmarks are closest in spirit to our goal of reproducible point-process
evaluation, but they operate primarily on temporal or marked temporal event
sequences rather than continuous spatiotemporal event data. \textsc{Seahorse}
fills this gap by standardizing data handling, model interfaces, training,
raw-space likelihood reporting, and diagnostics for heterogeneous STPP models
under a common experimental pipeline.

\section{Conclusions and Future Work}

We introduced \textsc{Seahorse}, a unified benchmarking framework for spatiotemporal point process models. It places heterogeneous model families under one experimental contract, so that comparisons no longer depend on each implementation's preprocessing, splits, training and reporting.

On the real datasets no model dominates: the best model changes with the dataset, the best-calibrated model is often not the likelihood winner, and a model that trains in minutes comes close to those that need hours. On the controlled suite only AutoSTPP stays below the separable reference as the coupling grows: a case for diagnostic benchmarks over leaderboards.

Three choices fixed here are open questions themselves: how many preceding events a model sees, whether events carry marks or covariates, and a single Hawkes generator. They limit the study, not the framework: with the groundwork in place, one can now ask how much history neural models need, how models transfer across regions and periods, and how they fare under other generators.

\begin{ack}
This work was supported by the German Federal Ministry of Education and Research (BMBF) under grant no.~01W23005 (project EVENTFUL). We thank Ismail Drief and Raphael Sonabend-Friend for their help in the early stages of the project, and the authors of the reference implementations on which \textsc{Seahorse} builds (Table~\ref{tab:reference-implementations}) for releasing their code. The authors declare no competing interests.
\end{ack}

\bibliographystyle{unsrt} 
\bibliography{references}
\appendix

\section{Software Interface and Reproducibility}
\label{app:software}
\begin{figure}
    \centering
    \includegraphics[width=\linewidth]{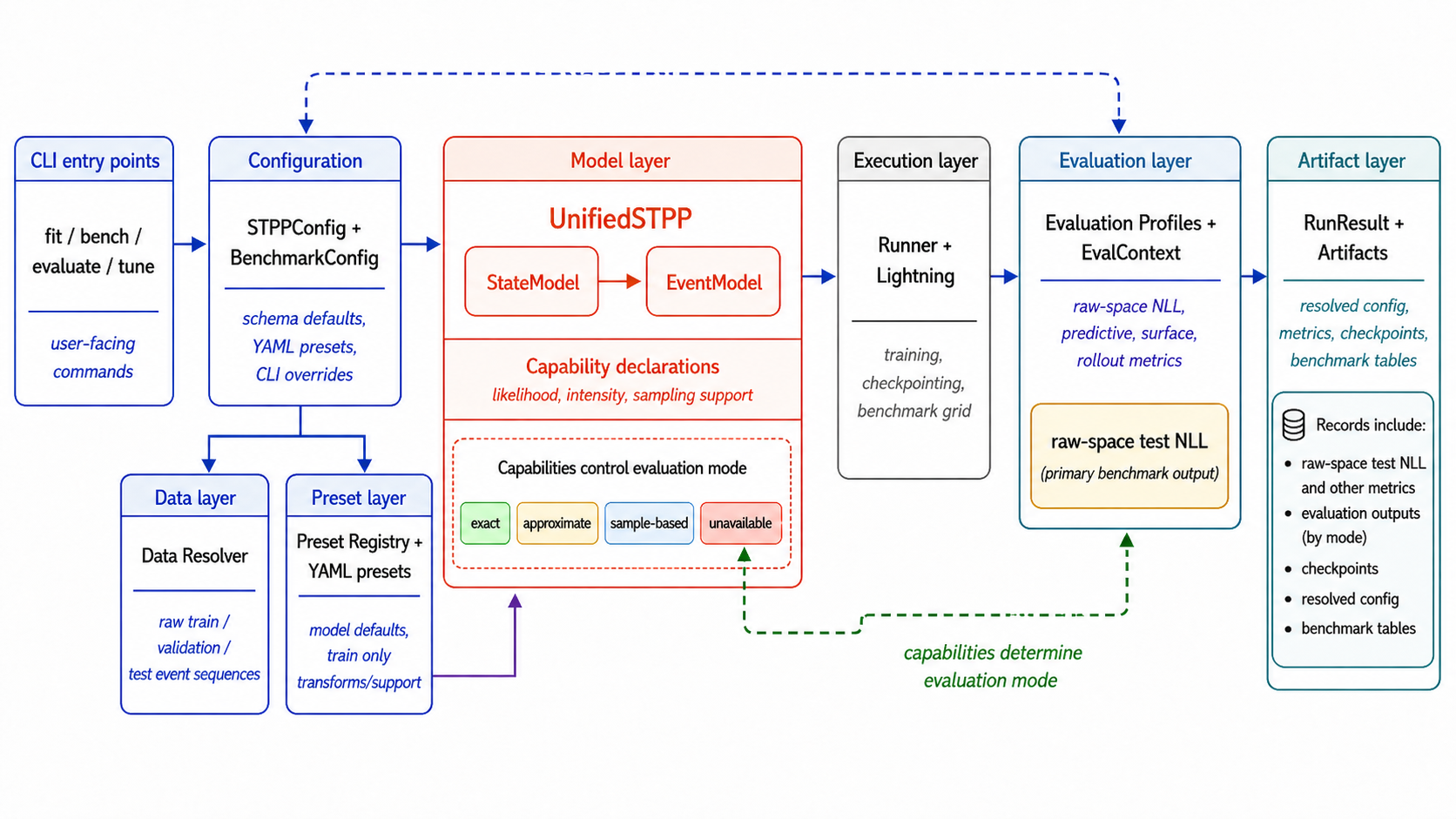}
\caption{
\textbf{\textsc{Seahorse} software architecture.}
The CLI resolves schema-validated configuration objects, dataset adapters expose raw event splits, preset registries construct \texttt{UnifiedSTPP} models, and runner/evaluation layers write structured artifacts. The architecture separates configuration, data resolution, model construction, execution, evaluation, and artifact recording.
}
        \label{fig:software-architecture}
\end{figure}
\subsection{Architecture Overview}
\label{app:software-architecture}
Figure~\ref{fig:software-architecture} summarizes the internal software architecture of \textsc{Seahorse}. The framework separates configuration, data handling, model construction, execution, evaluation, and artifact recording. A benchmark run begins with schema-validated configuration objects and fixed raw event splits. Dataset adapters expose event sequences in a common format, while model presets derive training-only quantities such as support bounds, coordinate transforms, and decoder-specific parameters. The resolved configuration is then used to instantiate the model components described in Section~\ref{sec:design-space}. Runners execute training, tuning, benchmark campaigns, or post-fit evaluation, and all outputs are written to structured artifacts that record the objective, reporting space, checkpoint rule, evaluation path, and preset status. \rev{The documentation and implementation are available at \repo.}

\subsection{Getting Started Guide}
\label{app:beginner-interface}
\paragraph{Run an Included Model.}
The simplest use case is to select an existing dataset split, choose a model preset, and launch a single training run. The user does not need to construct model objects manually: the preset resolves the required model components, training-only transforms, and reporting metadata.

\begin{codeblock}
python -m seahorse fit \
  --preset auto_stpp \
  --train data/example/train.jsonl \
  --val data/example/val.jsonl \
  --test data/example/test.jsonl \
  --out runs/auto_stpp_example
\end{codeblock}

The output directory contains the resolved configuration, selected checkpoint, metric logs, and a \textsc{RunResult}. This path is intended for users who want to evaluate one included STPP model on a provided or newly formatted event dataset.
\paragraph{Run a Benchmark Campaign.}
The benchmark interface launches a matrix of datasets, presets, and seeds under a frozen policy. This is the path used for the paper tables: each cell is executed under the same split, reporting-space convention, checkpoint rule, and status accounting.

\begin{codeblock}
python -m seahorse bench \
  --presets auto_stpp deep_stpp nsmpp \
  --splits_dir splits \
  --seeds 1 2 3 \
  --out runs/benchmark_example
\end{codeblock}

Benchmark campaigns write aggregate tables together with per-cell metadata. Failed, partial, missing, and still-running cells are represented explicitly rather than silently dropped.
\subsection{Researcher Path: Add or Wrap a Model}
\label{app:researcher-interface}

\textsc{Seahorse} is designed to make new STPP models immediately comparable to the benchmark suite. A researcher can add a method through either a native \textsc{Seahorse} implementation or a compatibility wrapper for an existing PyTorch/PyTorch Lightning model.

\paragraph{Native \textsc{Seahorse} implementation.}
If the model naturally fits the encode--evolve--decode abstraction, the preferred path is to implement the internal state/event decomposition. The researcher defines a \texttt{StateModel} for history encoding and state evolution, an \texttt{EventModel} for the conditional event law, and composes them as a \texttt{UnifiedSTPP}. The event model declares its supported benchmark queries through \texttt{EventCapabilities}: \rev{the likelihood type (exact or approximate) and its computation route, and whether the model exposes an intensity, a density, a score, or a native sampler}. The model is then registered as a named preset through \texttt{BaseModelConfig} and \texttt{ConfigRegistry}, with default hyperparameters stored in a preset YAML file. Once registered, the model can be launched with the same \texttt{fit}, \texttt{bench}, and \texttt{evaluate} commands used for the included baselines.

\begin{codeblock}
class MyStateModel(StateModel):
    def encode_history(self, times, locations, lengths, **kwargs):
        return StateContext(payload={"history_state": ...})

class MyEventModel(EventModel):
    @property
    def capabilities(self):
        return EventCapabilities(
            training_objective="nll",
            nll_kind="exact",
            has_intensity=True,
            has_native_sampler=True,
        )

    def training_loss(self, state_context, times, locations, lengths, **kwargs):
        return ...

@ConfigRegistry.register("my_stpp")
@dataclass
class MySTPPConfig(BaseModelConfig):
    def build_model(self):
        return UnifiedSTPP(
            state_model=MyStateModel(...),
            event_model=MyEventModel(...),
        )
\end{codeblock}

\paragraph{Compatibility wrapper for external models.}
Some STPP implementations do not naturally follow the \textsc{Seahorse} internal decomposition. In this case, the researcher can wrap an existing PyTorch or PyTorch Lightning model behind an \texttt{EventModel}-style adapter. The wrapper exposes the training loss and declares the subset of benchmark queries the model actually supports. For example, a model may support training and sampling but not exact raw-space likelihood or pointwise intensity evaluation. This limited exposure is not treated as an error: it is recorded through \texttt{EventCapabilities} and \textsc{RunResult} metadata, so benchmark tables can distinguish exact likelihoods, approximate likelihoods, sample-based diagnostics, and unavailable metrics.

\begin{codeblock}
class ExternalEventAdapter(EventModel):
    def __init__(self, external_model):
        super().__init__()
        self.external_model = external_model

    @property
    def capabilities(self):
        return EventCapabilities(
            training_objective="external_loss",
            nll_kind="approx",
            has_native_sampler=True,
            has_intensity=False,
        )

    def training_loss(self, state_context, times, locations, lengths, **kwargs):
        return self.external_model.loss(times, locations, lengths)
\end{codeblock}

Typical extension points are:
\begin{codeblock}
seahorse/models/state_models/,
seahorse/models/event_models/,
seahorse/models/configs/,
seahorse/configs/<preset>.yaml.
\end{codeblock}
This design lets new methods enter the same benchmark grid without rewriting data loaders, split handling, raw-space likelihood correction, metric computation, or result aggregation. It also prevents unsupported quantities from being silently fabricated: if a wrapped model cannot expose exact NLL, intensity surfaces, or \rev{samples}, the corresponding capability is marked unavailable and propagated into the reported artifacts.

\paragraph{\rev{Guides and tutorials.}}
\rev{The documentation at \docs{} contains a guide to extending the framework (wrapping a model, writing a preset, declaring capabilities, and testing a new model), a guide to adding and converting a dataset, and three tutorial notebooks that can be opened in Google Colab: fitting one model with the Python API, benchmarking models with the command-line interface, and a case study on the NYC collisions data. The repository's test suite includes tests of the preset and runner registries.}

\section{Model Implementations and Presets}
\label{app:models}

\subsection{Benchmark Model Summaries}
\label{app:benchmark_model_summaries}

\rev{Table~\ref{tab:model-aliases} lists the 19 benchmarked presets with their short aliases, preset identifiers and full names; Table~\ref{tab:unified-design-space} decomposes the neural methods.} The paragraphs below give implementation-level context for the presets used in our experiments and identify the source method or implementation followed where applicable.
\begin{table}[t]
\centering
\footnotesize
\setlength{\tabcolsep}{3pt}
\caption{Short aliases used in all result tables and figures, with the preset identifier and the full model name.}
\label{tab:model-aliases}
\begin{tabularx}{\linewidth}{l l Y}
\toprule
\textbf{Alias} & \textbf{Preset} & \textbf{Model} \\
\midrule
\multicolumn{3}{l}{\emph{Classical}} \\
HPP & \texttt{homogeneous\_poisson} & Homogeneous Poisson process (single spatial cell) \\
Seas-PP & \texttt{seasonal\_poisson} & Seasonal Poisson process (weekly phases $\times$ $8{\times}8$ spatial grid) \\
Spat-PP & \texttt{spatial\_poisson} & Spatial Poisson process ($8{\times}8$ spatial grid) \\
Hawkes-EG & \texttt{hawkes\_exp\_gaussian} & Hawkes process, exponential-in-time $\times$ Gaussian-in-space kernel \\
Gauss-PP & \texttt{gaussian\_poisson} & Gaussian Poisson process (homogeneous in time, Gaussian spatial background) \\
S-Hawkes & \texttt{static\_hawkes} & Static Hawkes process (Gaussian displacement independent of event age) \\
\midrule
\multicolumn{3}{l}{\emph{Factorized}} \\
Poisson & \texttt{poisson\_gmm} & Poisson temporal process + conditional GMM spatial head \\
Hawkes & \texttt{hawkes\_gmm} & Hawkes temporal process + conditional GMM spatial head \\
SC & \texttt{selfcorrecting\_gmm} & Self-correcting temporal process + conditional GMM spatial head \\
RMTPP & \texttt{rmtpp\_gmm} & RMTPP~\citep{du2016recurrent} + conditional GMM spatial head \\
THP & \texttt{thp\_gmm} & Transformer Hawkes process~\citep{zuo2020transformer} + conditional GMM spatial head \\
\midrule
\multicolumn{3}{l}{\emph{Neural STPP}} \\
AutoSTPP & \texttt{auto\_stpp} & AutoSTPP~\citep{zhou2024automatic} \\
DeepSTPP & \texttt{deep\_stpp} & DeepSTPP~\citep{zhou2022neural} \\
NSMPP & \texttt{nsmpp} & Neural spectral marked point process~\citep{zhu2022neural} \\
NJSDE & \texttt{njsde} & Neural jump SDE~\citep{jia2019neural} \\
Attn-CNF & \texttt{neural\_attncnf} & Neural STPP, attentive CNF~\citep{chen2021neural} \\
Jump-CNF & \texttt{neural\_jumpcnf} & Neural STPP, jump CNF~\citep{chen2021neural} \\
\midrule
\multicolumn{3}{l}{\emph{Generative}} \\
SMASH & \texttt{smash} & SMASH~\citep{li2024beyond} \\
DSTPP & \texttt{diffusion\_stpp} & Diffusion STPP~\citep{yuan2023spatio} \\
\bottomrule
\end{tabularx}
\end{table}

\paragraph{\rev{Classical baselines.}}
\rev{Six presets are classical statistical models: a homogeneous Poisson process with a single spatial cell (HPP); a spatial Poisson process on a grid (Spat-PP) and a seasonal one with weekly phases (Seas-PP), with $4\times4$, $8\times8$ or $16\times16$ cells chosen on validation data; a Hawkes process with an exponential-in-time and Gaussian-in-space kernel (Hawkes-EG); a Poisson process homogeneous in time with a Gaussian spatial background (Gauss-PP); and a static Hawkes process whose Gaussian displacement does not depend on the age of the event (S-Hawkes). HPP, Spat-PP, Seas-PP and Hawkes-EG are fitted by maximum likelihood without gradient training; Gauss-PP and S-Hawkes are trained like the neural presets (Table~\ref{tab:hpo_budget}).}

\paragraph{Factorized classical baselines.}
\rev{The factorized baselines combine a Poisson, Hawkes or self-correcting temporal process with a conditional Gaussian-mixture spatial head (\texttt{poisson\_gmm}, \texttt{hawkes\_gmm}, \texttt{selfcorrecting\_gmm}). They separate temporal-process assumptions from spatial-decoder expressivity. The release also provides the same temporal processes with time-invariant and time-varying continuous-normalizing-flow spatial heads (\texttt{*\_cnf}, \texttt{*\_tvcnf}), which are not benchmarked here.}

\paragraph{RMTPP and THP with spatial heads.}
RMTPP~\citep{du2016recurrent} and THP~\citep{zuo2020transformer} are temporal neural point-process baselines. In \textsc{Seahorse}, they are paired with a conditional Gaussian-mixture spatial head and reported as \texttt{rmtpp\_gmm} and \texttt{thp\_gmm}. These presets test whether strong temporal history encoders remain competitive when extended to continuous spatial prediction through a shared spatial decoder.

\paragraph{DeepSTPP.}
DeepSTPP~\citep{zhou2022neural} is implemented as a neural likelihood model for spatiotemporal event dynamics. It uses learned event-history representations and a flexible space--time intensity parameterization, providing a representative window-based neural STPP baseline in our benchmark.

\paragraph{AutoSTPP.}
AutoSTPP~\citep{zhou2024automatic} is an intensity-based STPP model built around automatic integration. Its integral network is parameterized so that differentiating the integrated quantity recovers the intensity, enabling exact likelihood evaluation with an exact compensator. In \textsc{Seahorse}, this preset represents joint-intensity models with explicit integration structure.

\paragraph{NSMPP.}
NSMPP~\citep{zhu2022neural} models the joint conditional intensity over time, space, and marks using learned spectral feature maps. In our implementation, it acts as a direct joint-intensity baseline with a numerical compensator.

\paragraph{NJSDE.}
NJSDE~\citep{jia2019neural} is a continuous-time latent-dynamics model based on neural jump stochastic differential equations. In \textsc{Seahorse}, the \texttt{njsde} preset uses this backbone with a conditional Gaussian-mixture spatial decoder, providing a continuous-time neural baseline with explicit event-driven state evolution.

\paragraph{Neural STPP variants.}
The Neural STPP family~\citep{chen2021neural} uses continuous-time hidden dynamics for event histories and conditional normalizing flows for spatial likelihoods. We include two variants: \texttt{neural\_jumpcnf}, with a jump-conditioned continuous normalizing flow, and \texttt{neural\_attncnf}, with an attention-conditioned continuous normalizing flow. These presets test the effect of flow-based spatial likelihoods and continuous-time state evolution under the shared benchmark contract.
\paragraph{SMASH.}
SMASH~\citep{li2024beyond} is a score-matching-based generative STPP model. It replaces exact likelihood optimization with a normalization-free pseudolikelihood objective and generates event samples through score-based dynamics. In \textsc{Seahorse}, SMASH represents sample-based generative models whose likelihood and intensity diagnostics require explicit evaluation-path metadata.

\paragraph{DSTPP.}
DSTPP~\citep{yuan2023spatio} models future spatiotemporal events with a diffusion-based generative process. Rather than exposing a conventional conditional intensity, it generates next-event samples through denoising dynamics conditioned on event history. This preset provides a second sample-based generative family and motivates the benchmark distinction between exact likelihoods, approximate likelihoods, and sampling-based diagnostics.

Where available, our implementations follow the public reference repositories released by the original authors; repository links and exact preset configurations are included in the released code documentation.
\subsection{Reference Implementations}
\label{app:reference-implementations}

Where available, \textsc{Seahorse} presets follow the public reference implementations released by the original authors. Table~\ref{tab:reference-implementations} lists the implementation sources used for alignment and cross-checking.

\begin{table}[t]
\centering
\small
\caption{Reference implementations used for preset alignment.}
\label{tab:reference-implementations}
\begin{tabularx}{\linewidth}{l X}
\toprule
\textbf{Preset family} & \textbf{Reference implementation} \\
\midrule
DeepSTPP &
\href{https://github.com/Rose-STL-Lab/DeepSTPP}{\texttt{Rose-STL-Lab/DeepSTPP}} \\

AutoSTPP &
\href{https://github.com/Rose-STL-Lab/AutoSTPP}{\texttt{Rose-STL-Lab/AutoSTPP}} \\

Neural AttnCNF &
\href{https://github.com/facebookresearch/neural_stpp}{\texttt{facebookresearch/neural\_stpp}} \\
Neural JumpCNF &
\href{https://github.com/facebookresearch/neural_stpp}{\texttt{facebookresearch/neural\_stpp}} \\
NJSDE &
\href{https://github.com/facebookresearch/neural_stpp}{\texttt{facebookresearch/neural\_stpp}} \\
NSMPP &
\leavevmode\rev{\href{https://github.com/meowoodie/Neural-Spectral-Marked-Point-Processes}{\texttt{meowoodie/Neural-Spectral-Marked-Point-Processes}}} \\
SMASH &
\href{https://github.com/zichongli5/SMASH}{\texttt{zichongli5/SMASH}} \\

DSTPP &
\href{https://github.com/tsinghua-fib-lab/Spatio-temporal-Diffusion-Point-Processes}
{\texttt{tsinghua-fib-lab/Spatio-temporal-Diffusion-Point-Processes}} \\
factorized &
\href{https://github.com/facebookresearch/neural_stpp}{\texttt{facebookresearch/neural\_stpp}} \\
RMTPP &
\href{https://github.com/Rose-STL-Lab/AutoSTPP}{\texttt{Rose-STL-Lab/AutoSTPP}} \\
THP &
\href{https://github.com/SimiaoZuo/Transformer-Hawkes-Process}{\texttt{SimiaoZuo/Transformer-Hawkes-Process}} \\
\bottomrule
\end{tabularx}
\end{table}

\section{Evaluation Semantics and Metrics}
\label{app:evaluation}

\subsection{Raw-Space Held-Out Likelihood}
\label{app:ll-computation}

All likelihood-based evaluations target the same event-wise object: the conditional density of each observed event given its pre-event history,
\[
    \rev{p(t_i,\mathbf{s}_i \mid \mathcal{H}_{t_i})} .
\]
The per-event negative log-likelihood is
\begin{equation}
    \mathrm{NLL}_i
    =
    -\log \rev{p(t_i,\mathbf{s}_i \mid \mathcal{H}_{t_i})},
    \label{eq:per-event-nll}
\end{equation}
and the reported test NLL is the average over test events,
\[
    \mathrm{NLL}
    =
    \frac{1}{N}\sum_{i=1}^{N}\mathrm{NLL}_i .
\]
For intensity-based models, the same quantity is evaluated through the equivalent event term and compensator form. Score-based and diffusion models do not generally expose exact likelihoods; when they appear in likelihood tables, their evaluation path is marked as approximate or variational.

\paragraph{Coordinate correction.}
Many STPP models rescale time and space internally for numerical stability. Let \(z=\rev{(\Delta t,\mathbf{s})}\) denote raw coordinates and \(u=g(z)\) the model's native coordinates. If \(g\) is affine with Jacobian
\[
    \left|\det \frac{\partial g}{\partial z}\right|
    =
    \prod_d r_d^{-1},
\]
then raw-space and native-space NLLs satisfy
\begin{equation}
    \mathrm{NLL}^{(i)}_{\mathrm{raw}}
    =
    \mathrm{NLL}^{(i)}_{\mathrm{native}}
    +
    \sum_d \log r_d .
    \label{eq:jacobian-correction}
\end{equation}
All benchmark-facing NLL values are reported in raw coordinates after applying the corresponding correction. This avoids comparing likelihoods defined with respect to different coordinate measures.

\subsection{Objective, Capability, and Diagnostic Semantics}
\label{app:evaluation-semantics}

The optimized training objective is not always the benchmark quantity reported in the tables. Some models optimize exact point-process likelihoods, others optimize density likelihoods, numerical likelihoods, variational objectives, or score-matching losses. \textsc{Seahorse} therefore records the likelihood evaluation path in the run metadata, rather than treating all scalar losses as comparable NLLs.

The benchmark is also capability-aware. It does not assume that every model exposes the same queries: some models provide pointwise intensities, some provide normalized conditional densities, and some provide only samples. These differences determine which diagnostics can be computed directly. \textsc{Seahorse} records whether a fitted model supports likelihood evaluation, intensity queries, sampling, \rev{and} surface evaluation; unsupported quantities are marked unavailable rather than inferred implicitly.

Beyond the metrics reported in the main text, the implementation makes a larger diagnostic catalog available for model inspection. Teacher-forced metrics condition on the realized pre-event history and evaluate the next observed event, isolating one-step predictive quality from autoregressive error accumulation. Surface-recovery metrics apply to synthetic suites with known ground-truth intensity. PIT metrics are reported as deviations from the uniform calibration target, so lower values indicate better calibrated predictive distributions. The paper reports only the subset most relevant to likelihood comparability, teacher-forced prediction\rev{, calibration}, \rev{and} intensity-surface recovery.

\begin{table}[t]
\centering
\small
\caption{NLL evaluation paths. ``Training loss'' is the optimized objective; ``Eval NLL'' is the reported benchmark likelihood path.}
\label{tab:nll-paths}
\begin{tabularx}{\linewidth}{l X X X}
\toprule
\textbf{Model family} & \textbf{Training loss} & \textbf{Eval NLL} & \textbf{Annotation} \\
\midrule
AutoSTPP & Point-process NLL & Exact compensator via integral network & exact \\
DeepSTPP & Model-native likelihood / variational objective & Evaluated density or likelihood path & \leavevmode\rev{exact} \\
NSMPP & Point-process likelihood & Numerical compensator & \leavevmode\rev{quadrature} \\
NJSDE / Neural STPP variants & Likelihood with auxiliary regularization & \leavevmode\rev{Likelihood integrated by an adaptive ODE solver} & \leavevmode\rev{ODE solver} \\
RMTPP / THP & Temporal likelihood with spatial density head & Factorized next-event likelihood & \leavevmode\rev{exact / quadrature} \\
DSTPP & Diffusion denoising / variational objective & \leavevmode\rev{Full variational bound (100 diffusion steps), an upper bound on the NLL} & \leavevmode\rev{bound} \\
SMASH & Score-matching pseudolikelihood & \leavevmode\rev{Temporal NLL plus the spatial density recovered from the score on a $128\times128$ grid} & \leavevmode\rev{approximation} \\
\bottomrule
\end{tabularx}
\end{table}

\begin{table*}[t]
\centering
\small
\caption{Evaluation metrics used in \textsc{Seahorse}. Not all metrics apply to all model families.}
\label{tab:metric-catalog}
\begin{tabularx}{\textwidth}{l l c X}
\toprule
\textbf{Metric} & \textbf{Regime} & \textbf{Direction} & \textbf{Purpose} \\
\midrule
NLL & Held-out likelihood & \(\downarrow\) & Raw-coordinate event likelihood fit \\
Temporal CRPS & Teacher-forced prediction & \(\downarrow\) & Distributional next-event time prediction \\
Spatial energy score & Teacher-forced prediction & \(\downarrow\) & Spatial predictive distribution quality \\
Temporal PIT error & Calibration & \(\downarrow\) & Deviation from the uniform temporal PIT target \\
Spatial PIT error & Calibration & \(\downarrow\) & Deviation from the spatial calibration target \\
Temporal MAE & Point prediction & \(\downarrow\) & Error of temporal point summary \\
Spatial MAE / RMSE & Point prediction & \(\downarrow\) & Spatial point-summary error \\
Joint distance & Teacher-forced prediction & \(\downarrow\) & Joint time-space next-event error \\
Intensity correlation & Synthetic surface recovery & \(\uparrow\) & Correlation with ground-truth intensity surface \\
Log-intensity RMSE & Synthetic surface recovery & \(\downarrow\) & Magnitude error on log-intensity scale \\
Mass placement error & Synthetic surface recovery & \(\downarrow\) & Misplacement of predicted mass relative to ground-truth intensity \\
\bottomrule
\end{tabularx}
\end{table*}

\section{Datasets and Benchmark Protocols}
\label{app:datasets}

\subsection{Dataset Summary}
\label{app:dataset-summary}

\rev{Table~\ref{tab:dataset-summary} lists the thirteen real datasets with their splits, licences and availability. All are published on the Hugging Face Hub under \href{https://huggingface.co/seahorse-stpp}{\texttt{huggingface.co/seahorse-stpp}} with a dataset card and Croissant metadata. The Citi Bike data (both versions) are not redistributed because the publisher's licence does not allow it, and Seattle~911 is not hosted; for these three, the repositories contain scripts that rebuild the data from the original source. The COVID data are \copyright{} The New York Times, for non-commercial use only, with credit to The New York Times.}

\begin{table*}[t]
\centering
\footnotesize
\setlength{\tabcolsep}{3pt}
\caption{\rev{Real datasets. Period: time span; split points: start of the validation and of the test period (chronological splits); original: the original splits of the Neural STPP benchmark~\citep{chen2021neural}, which overlap in time (App.~\ref{sec:table-notes}). Hosted: data on the Hugging Face Hub; script: rebuild script only.}}
\label{tab:dataset-summary}
\begin{tabularx}{\textwidth}{l >{\raggedright\arraybackslash}X l l l}
\toprule
\rev{Dataset} & \rev{Events, region} & \rev{Period; split points} & \rev{Licence} & \rev{Hosted} \\
\midrule
\rev{ComCat} & \rev{earthquakes, California (USGS ComCat)} & \rev{1981--2020; split at 1998, 2007} & \rev{public domain (USGS)} & \rev{yes} \\
\rev{GeoNet} & \rev{earthquakes, New Zealand (GeoNet)} & \rev{2018--2020; split at 2019-01, 2019-07} & \rev{CC BY 3.0 NZ} & \rev{yes} \\
\rev{Citi Bike (h)} & \rev{bike-share trip starts, New York City; hourly sequences} & \rev{2019-04--08; split at 08-01, 08-16} & \rev{publisher licence} & \rev{script} \\
\rev{Austin} & \rev{311 service requests, Austin} & \rev{2023; split at 09-14, 10-20} & \rev{public domain (City of Austin)} & \rev{yes} \\
\rev{Seattle} & \rev{fire-department 911 dispatches, Seattle} & \rev{2023, as Austin} & \rev{public domain (City of Seattle)} & \rev{script} \\
\rev{NYC 311} & \rev{311 service requests, New York City} & \rev{2023, as Austin} & \rev{NYC Open Data terms} & \rev{yes} \\
\rev{NYC coll.} & \rev{traffic collisions, New York City} & \rev{2023, as Austin} & \rev{NYC Open Data terms} & \rev{yes} \\
\rev{Chicago} & \rev{crime reports, Chicago} & \rev{2023, as Austin} & \rev{City of Chicago terms} & \rev{yes} \\
\rev{LA} & \rev{crime reports, Los Angeles} & \rev{2023, as Austin} & \rev{CC0} & \rev{yes} \\
\rev{Uber} & \rev{ride-hailing pickups, New York City} & \rev{2014-04--09; split at 08-01, 09-01} & \rev{CC0} & \rev{yes} \\
\midrule
\rev{COVID} & \rev{COVID-19 case reports, New Jersey counties (NYT)} & \rev{2020-03--07; original} & \rev{NYT, non-commercial} & \rev{yes} \\
\rev{Earthquakes} & \rev{earthquakes $M\ge2.5$, Japan (USGS ComCat)} & \rev{1990--2019; original} & \rev{public domain (USGS)} & \rev{yes} \\
\rev{Citi Bike (d)} & \rev{bike-share trip starts, New York City; daily sequences} & \rev{2019-04--08; original} & \rev{publisher licence} & \rev{script} \\
\bottomrule
\end{tabularx}
\end{table*}

\subsection{Raw Event Splits and Preprocessing}
\label{app:dataset-protocol}

\rev{The ten benchmark datasets are split chronologically into training, validation and test periods (Table~\ref{tab:dataset-summary}), and each period is cut into sequences of one day (one hour for Citi Bike (h) and Uber). Every model is scored on the same targets, each conditioned on its 20 preceding events, and likelihoods are reported in raw units (time in days, space in degrees). The three Neural STPP datasets keep their original splits into independent windows (App.~\ref{sec:table-notes}).} Model-specific normalization, support bounds, and coordinate transforms are estimated from the training split only and stored with the run artifacts. Reported likelihoods are mapped back to raw coordinate space as described in Appendix~\ref{app:ll-computation}.

\subsection{Controlled Synthetic Suites}
\label{app:synthetic-datasets}

\rev{The HawkesNest entanglement suite~\citep{aalaila2026hawkesnest} varies one structural axis, the coupling between the time and the location of triggered events, while holding the background and the branching ratio fixed, and stores the true intensity for intensity-recovery diagnostics. Its specification follows; results are in Section~\ref{sec:experiments} and Appendix~\ref{app:additional-learning-curves}.}

\paragraph{\rev{Generator.}}\label{app:hawkesnest-generator}
\rev{The entanglement suite is a single-mark spatiotemporal Hawkes process on the unit square $\mathcal{S}=[0,1]^2$ with conditional intensity
\begin{equation}\label{eq:hawkesnest-intensity}
\lambda^{\ast}(t,\mathbf{s}\mid\mathcal{H}_t)
= \mu(\mathbf{s})
+ a\sum_{\substack{t_i<t\\ t-t_i\le\tau_{\max}}}\varphi_v\big(t-t_i,\ \mathbf{s}-\mathbf{s}_i\big),
\qquad \tau_{\max}=5 .
\end{equation}
The background is constant in time; its logarithm is a constant plus three Gaussian bumps,
\begin{equation}
\mu(\mathbf{s})=\exp\!\Big(a_0+A\sum_{k=1}^{3}\exp\!\big(-\|\mathbf{s}-\mathbf{c}_k\|^2/(2\sigma_b^2)\big)\Big),
\quad a_0=-0.2997,\ A=2.5,\ \sigma_b=0.20,
\end{equation}
with centres $\mathbf{c}_k\in\{(0.25,0.25),(0.75,0.25),(0.75,0.75)\}$; $a_0$ is set so that $\mu$ integrates to $4.35$ events per time unit over $\mathcal{S}$, $30\%$ of a target rate of $14.5$. The triggering kernel has unit amplitude, an exponential time profile and a Gaussian spatial profile whose centre travels at speed $v$ in the direction $\hat{\mathbf{e}}=(\cos\theta,\sin\theta)$, $\theta=\pi/4$:
\begin{equation}\label{eq:hawkesnest-kernel}
\varphi_v(\tau,\mathbf{r})=\exp(-\tau/\beta)\,\exp\!\Big(-\frac{\|\mathbf{r}-v\tau\,\hat{\mathbf{e}}\|^2}{2\sigma^2}\Big),
\qquad \beta=0.3,\ \sigma=0.03 .
\end{equation}
The levels $\mathbf{L}_0$--$\mathbf{L}_3$ set $v=0,0.1,0.2,0.3$. We measure entanglement by $\rho=v\beta/\sigma=0,1,2,3$, the distance the offspring cloud drifts during one temporal scale $\beta$, in units of its width $\sigma$. At $\rho=0$ the kernel is separable in time and space; for $\rho>0$ the spatial law of an offspring depends on its age. Only $v$ and $a$ change between levels.}

\paragraph{\rev{Branching and background share.}}
\rev{The kernel mass $\int_0^{\tau_{\max}}\!\!\int_{\mathbb{R}^2}\varphi_v\,d\mathbf{r}\,d\tau=2\pi\sigma^2\beta(1-e^{-\tau_{\max}/\beta})$ does not depend on $v$, so $a=\eta/\big(2\pi\sigma^2\beta(1-e^{-\tau_{\max}/\beta})\big)$ for a nominal branching ratio $\eta$. Offspring that fall outside $\mathcal{S}$ are discarded, and a drifting cloud loses more of them, so the nominal $\eta$ is raised with the level ($0.712$, $0.729$, $0.757$, $0.793$) to keep the realized in-square branching ratio at about $0.70$. On the generated data, at every level, the realized branching ratio (in-square kernel mass per event) is $0.700$--$0.703$, the rate is $14.56$--$14.64$ events per time unit over all windows ($14.51$--$14.57$ averaged over realizations), the triggered share of events, $1-\int\mu/\text{rate}$ with the window rate, is $0.701$--$0.703$, and the background accounts for $29$--$30\%$ of $\lambda^\ast$ at the events.}

\paragraph{\rev{Simulation.}}
\rev{Events are simulated with the cluster (immigrant--offspring) representation: immigrants form a Poisson process with intensity $\mu$ on $[-20,700]$ (20 time units of burn-in); every event has a Poisson$(\eta)$ number of offspring, with ages drawn from the exponential law truncated at $\tau_{\max}$ and displacements from $\mathcal{N}(v\tau\hat{\mathbf{e}},\sigma^2 I)$; offspring outside $\mathcal{S}$ are dropped, which gives the same law as thinning with the kernel restricted to $\mathcal{S}$. Events at $t\ge0$ are kept and each realization is truncated to its first $8{,}000$ events.}

\paragraph{\rev{Realizations, windows and splits.}}
\rev{Each level has ten realizations of $8{,}000$ events, spanning $514$--$578$ time units. Realization $r$ uses the generator seed $20261102+r$ at every level, so the immigrants are identical across levels (common random numbers). Each realization is cut into $20$ consecutive windows of equal duration ($25.7$--$28.9$ time units, $235$--$587$ events). For realizations $0$--$4$, windows $0$--$15$ form the training set, $16$--$17$ the validation set and $18$--$19$ the test set; realizations $5$--$9$ contribute only windows $18$--$19$, to the test set. Every level therefore has $80/10/20$ training, validation and test sequences, with the test windows at the end of their realizations. Times are stored relative to the window start and are strictly increasing in single precision (one tie per level was separated by one single-precision ulp; recorded in the data manifest). Each window is an independent sequence: the events of earlier windows are not shown to the models. All presets are scored on the common target set of events with index $\ge 20$ in their window, the history length of the windowed presets (AutoSTPP, DeepSTPP). Table~\ref{tab:hawkesnest-levels} lists the per-level parameters and counts.}

\begin{table}[h]
\centering\footnotesize
\setlength{\tabcolsep}{3pt}
\caption{\rev{HawkesNest entanglement suite: level parameters and counts. Sequences are windows; events are summed over the windows of a split; common targets are the test events with window index $\ge20$. Realized $\eta$: in-square kernel mass per event.}}
\label{tab:hawkesnest-levels}
\begin{tabular}{lccccccc}
\toprule
\rev{Level} & \rev{$v$} & \rev{$\rho$} & \rev{$\eta$ (nominal / realized)} & \rev{$a$} & \rev{train seq.\ / events} & \rev{val.\ seq.\ / events} & \rev{test seq.\ / events / targets} \\
\midrule
\rev{$\mathbf{L}_0$} & \rev{$0$}   & \rev{$0$} & \rev{$0.712$ / $0.700$} & \rev{$419.70$} & \rev{$80$ / $31{,}975$} & \rev{$10$ / $3{,}925$} & \rev{$20$ / $8{,}229$ / $7{,}829$} \\
\rev{$\mathbf{L}_1$} & \rev{$0.1$} & \rev{$1$} & \rev{$0.729$ / $0.703$} & \rev{$429.72$} & \rev{$80$ / $31{,}692$} & \rev{$10$ / $4{,}203$} & \rev{$20$ / $8{,}354$ / $7{,}954$} \\
\rev{$\mathbf{L}_2$} & \rev{$0.2$} & \rev{$2$} & \rev{$0.757$ / $0.701$} & \rev{$446.22$} & \rev{$80$ / $32{,}122$} & \rev{$10$ / $3{,}807$} & \rev{$20$ / $8{,}294$ / $7{,}894$} \\
\rev{$\mathbf{L}_3$} & \rev{$0.3$} & \rev{$3$} & \rev{$0.793$ / $0.701$} & \rev{$467.44$} & \rev{$80$ / $31{,}723$} & \rev{$10$ / $3{,}893$} & \rev{$20$ / $8{,}451$ / $8{,}051$} \\
\bottomrule
\end{tabular}
\end{table}

\paragraph{\rev{Ground-truth intensity.}}
\rev{For the intensity-recovery diagnostic, $\lambda^{\ast}$ of Eq.~\eqref{eq:hawkesnest-intensity} is evaluated from the true parameters on a $32\times32\times50$ space--time grid over every test window, conditioning on the full realization history, including the events before the window that the models do not observe.}

\paragraph{\rev{Excess over the true process and the separable oracle.}}
\rev{The NLL of the true process on the common targets changes with the level ($-3.168$, $-3.224$, $-3.247$, $-3.337$ nats/event at $\mathbf{L}_0$--$\mathbf{L}_3$), so raw NLLs are not comparable across levels. We therefore report the \emph{excess}: a model's NLL minus the NLL of $\lambda^\ast$ on the same targets, with $\lambda^\ast$ conditioned on the same window history that the models see. As a reference we fit the best separable Hawkes process, $c\,\mu(\mathbf{s})+\eta\sum_{t_i<t}\beta'^{-1}e^{-(t-t_i)/\beta'}\,\mathcal{N}(\mathbf{s}-\mathbf{s}_i;\mathbf{m},\Sigma)$, with the true background shape and free $c,\eta,\beta',\mathbf{m}$ and full covariance $\Sigma$, by maximum likelihood on the training windows. Its excess on the common test targets is $0.000$, $0.079$, $0.173$ and $0.250$ nats/event at $\mathbf{L}_0$--$\mathbf{L}_3$ (standard errors over test windows $\le0.008$): this is the likelihood lost by the best separable fit in this family (exponential in time, Gaussian in space). The separable kernel built from the two marginals of the true kernel loses more ($0.088$, $0.196$ and $0.302$ at $\mathbf{L}_1$--$\mathbf{L}_3$); other spatial profiles were not searched, so the reference is an upper bound on the loss of the best separable kernel. A maximum-likelihood refit of the true travelling-wave family on the same training windows recovers the true NLL to within $3\times10^{-4}$ nats/event at every level, so the gap of the separable fit is not an artifact of fitting from finite data. In the submitted version the suite used a weaker configuration of the same generator ($\sigma=0.15$, $\eta=0.40$, $v=0,0.05,0.15,0.30$, five realizations per level); there the best separable fit lost at most $0.002$ nats/event to the true intensity at any level, so the levels could not be told apart by likelihood.}

\paragraph{\rev{Data.}}
\rev{The suite (levels $\mathbf{L}_0$--$\mathbf{L}_3$), including the true-process parameters, the generator code and SHA-256 checksums of all files, is available under CC BY 4.0 at \href{https://huggingface.co/datasets/seahorse-stpp/seahorse_synthetic_stpp}{\texttt{huggingface.co/datasets/seahorse-stpp/seahorse\_synthetic\_stpp}} (folder \texttt{hawkesnest\_entanglement/}).}

\paragraph{Relation to other synthetic generators, and generator alignment.}
There is no agreed general-purpose synthetic data-generating process for neural STPPs; published work uses Poisson, log-Gaussian Cox, Hawkes or task-specific constructions. The Pinwheel suite of the Neural STPP paper~\citep{chen2021neural}, for instance, is itself a multivariate Hawkes process whose ten spatial clusters trigger one another clockwise. HawkesNest keeps the Hawkes mechanism because it is the construction under which the latent intensity, the branching structure and the entanglement are all available in closed form, which is what makes failure attribution possible; its modular formulation also allows the triggering to be switched off ($\eta=0$) and the background to be driven by richer mechanisms (moving or oscillating hotspots, travelling Gabor patterns, shear and vortex fields, regime switching), which we do not use here. A Hawkes generator favours models whose inductive bias matches it, so the absolute ranking on this suite must not be read as generator-independent; the suite is meant to diagnose how models degrade and where metrics disagree as the coupling increases, as in Section~\rev{\ref{sec:experiments}}. Fully non-classical generators are a natural extension and are left to future work.

\section{Additional Results}
\label{app:additional-results}

\rev{This section gives the full real-data NLL table, the rank overview, the calibration and next-event tables on all datasets, the HawkesNest calibration, and the tuning and compute records. All tables and figures are generated by scripts from the run artifacts.}
\begin{sidewaystable*}
\centering
\caption{\rev{Full version of Table~\ref{tab:real_data_expanded_nll}, with all per-cell markers (App.~\ref{sec:table-notes}).}}
\label{tab:real_data_full_nll}
\footnotesize
\setlength{\tabcolsep}{1pt}
\resizebox{\linewidth}{!}{%
\begin{tabular}{lccccccccccccc}
\toprule
\textbf{Preset} & \textbf{COVID} & \textbf{Earthquakes} & \textbf{Citi Bike (d)} & \textbf{ComCat} & \textbf{GeoNet} & \textbf{Citi Bike (h)} & \textbf{Austin} & \textbf{Seattle} & \textbf{NYC 311} & \textbf{NYC coll.} & \textbf{Chicago} & \textbf{LA} & \textbf{Uber} \\
\midrule
\multicolumn{14}{l}{\emph{Classical}} \\
HPP & $-0.193_{\pm0.00}$ & $6.405_{\pm0.00}$ & $-5.257_{\pm0.00}$ & $5.155_{\pm0.00}$ & $3.823_{\pm0.00}$ & $-11.326_{\pm0.00}$ & $-6.537_{\pm0.00}$ & $-7.100_{\pm0.00}$ & $-9.227_{\pm0.00}$ & $-5.627_{\pm0.00}$ & $-6.764_{\pm0.00}$ & $-6.241_{\pm0.00}$ & $-10.476_{\pm0.00}$ \\
Seas-PP & $-2.063_{\pm0.00}$ & $5.096_{\pm0.00}$ & $-8.752_{\pm0.00}$ & $3.576_{\pm0.00}$ & $0.863_{\pm0.00}$ & $-14.804_{\pm0.00}$ & $-8.307_{\pm0.00}$ & $-8.944_{\pm0.00}$ & $-10.854_{\pm0.00}$ & $-7.081_{\pm0.00}$ & $-8.384_{\pm0.00}$ & $-7.806_{\pm0.00}$ & $-13.529_{\pm0.00}$ \\
Spat-PP & $-2.056_{\pm0.00}$ & $5.093_{\pm0.00}$ & $-8.752_{\pm0.00}$ & $3.573_{\pm0.00}$ & $0.862_{\pm0.00}$ & $-14.805_{\pm0.00}$ & $-8.290_{\pm0.00}$ & $-8.943_{\pm0.00}$ & $-10.850_{\pm0.00}$ & $-7.081_{\pm0.00}$ & $-8.384_{\pm0.00}$ & $-7.805_{\pm0.00}$ & $-13.530_{\pm0.00}$ \\
Hawkes-EG & $-2.070_{\pm0.00}$ & $4.378_{\pm0.00}$ & $-8.813_{\pm0.00}$ & $0.796_{\pm0.00}$ & $0.980_{\pm0.00}$ & $-14.946_{\pm0.00}$ & $-8.400_{\pm0.00}$ & $-8.439_{\pm0.00}$ & $-10.609_{\pm0.00}$ & $-6.710_{\pm0.00}$ & $-7.839_{\pm0.00}$ & $-7.608_{\pm0.00}$ & $-13.484_{\pm0.00}$ \\
Gauss-PP & $-1.238_{\pm0.04}$ & $6.231_{\pm0.00}$ & $-8.718_{\pm0.00}$ & $4.557_{\pm0.01}$ & $1.849_{\pm0.01}$ & $-14.740_{\pm0.00}$ & $-8.118_{\pm0.01}$ & $-8.444_{\pm0.00}$ & $-10.353_{\pm0.00}$ & $-6.629_{\pm0.00}$ & $-7.981_{\pm0.00}$ & $-7.081_{\pm0.00}$ & $-12.724_{\pm0.01}$ \\
S-Hawkes & $-2.143_{\pm0.02}$ & $4.558_{\pm0.01}$ & $-8.910_{\pm0.00}$ & $0.579_{\pm0.00}$ & $0.940_{\pm0.00}$ & $-15.073_{\pm0.00}$ & $-8.556_{\pm0.00}$ & $-8.691_{\pm0.00}$ & $-11.569_{\pm0.04}$ & $-6.767_{\pm0.00}$ & $-8.480_{\pm0.01}$ & $-7.938_{\pm0.00}$ & $-13.610_{\pm0.00}$ \\
\midrule
\multicolumn{14}{l}{\emph{Factorized}} \\
Poisson & $-1.471_{\pm0.01}$ & $5.374_{\pm0.00}$ & $-8.707_{\pm0.00}$ & $2.979_{\pm0.00}$ & $1.342_{\pm0.00}$ & $-12.753_{\pm0.00}$ & $-7.744_{\pm0.00}$ & $-8.209_{\pm0.00}$ & $-7.502_{\pm0.00}$ & $-6.583_{\pm0.00}$ & $-7.327_{\pm0.00}$ & $-6.744_{\pm0.00}$ & $-11.649_{\pm0.00}$ \\
Hawkes & $-1.556_{\pm0.01}$ & $5.188_{\pm0.00}$ & $-8.861_{\pm0.00}$ & $2.302_{\pm0.01}$ & $1.327_{\pm0.00}$ & $-14.883_{\pm0.00}$ & $-8.490_{\pm0.00}$ & $-8.443_{\pm0.00}$ & $-10.143_{\pm0.00}$ & $-6.717_{\pm0.00}$ & $-8.114_{\pm0.00}$ & $-7.414_{\pm0.00}$ & $-13.008_{\pm0.01}$ \\
SC & $3.020_{\pm0.05}$ & $6.614_{\pm0.01}$ & $-7.094_{\pm0.02}$ & $3.083_{\pm0.00}$ & $1.848_{\pm0.01}$ & $-9.061_{\pm0.00}$ & $-4.669_{\pm0.01}$ & $-7.018_{\pm0.01}$ & $-4.207_{\pm0.01}$ & $-4.895_{\pm0.00}$ & $-5.579_{\pm0.01}$ & $-4.999_{\pm0.00}$ & $-8.107_{\pm0.00}$ \\
RMTPP & $-1.841_{\pm0.03}$ & $5.331_{\pm0.00}$ & $-8.785_{\pm0.00}$ & $2.229_{\pm0.01}$ & $1.335_{\pm0.00}$ & $-14.903_{\pm0.00}$ & $-8.336_{\pm0.00}$ & $-8.425_{\pm0.00}$ & $-10.432_{\pm0.00}$ & $-6.686_{\pm0.00}$ & $-8.165_{\pm0.00}$ & $-7.532_{\pm0.00}$ & $-12.943_{\pm0.00}$ \\
THP$^{\mathrm{q}}$ & $-2.012_{\pm0.01}$ & $5.117_{\pm0.00}$ & $-8.846_{\pm0.00}$ & $2.206_{\pm0.01}$ & $1.331_{\pm0.00}$ & $-15.048_{\pm0.00}$ & $-8.498_{\pm0.00}$ & $-8.444_{\pm0.00}$ & $-10.504_{\pm0.01}$ & $-6.710_{\pm0.00}$ & $-8.180_{\pm0.00}$ & $-7.433_{\pm0.03}$ & $-13.081_{\pm0.01}$ \\
\midrule
\multicolumn{14}{l}{\emph{Neural STPP}} \\
AutoSTPP & $-2.128_{\pm0.02}$ & \underline{$3.655_{\pm0.01}$} & $-8.913_{\pm0.02}$ & $\mathbf{-0.469_{\pm0.06}}$ & $\mathbf{0.315_{\pm0.14}}$ & $\mathbf{-15.717_{\pm0.20}}$ & \underline{$-8.873_{\pm0.03}$} & $-8.678_{\pm0.03}$ & $\mathbf{-12.203_{\pm0.48}}$ & $-6.796_{\pm0.01}$ & $-8.649_{\pm0.04}$ & $-8.291_{\pm0.06}$ & $-13.730_{\pm0.01}$ \\
DeepSTPP & $-2.391_{\pm0.08}$ & $\mathbf{3.411_{\pm0.03}}$ & $-9.214_{\pm0.02}$ & \underline{$-0.004_{\pm0.06}$} & $0.421_{\pm0.02}$ & $-15.433_{\pm0.01}$ & $\mathbf{-8.941_{\pm0.01}}$ & $-9.085_{\pm0.04}$ & \underline{$-11.573_{\pm0.01}$} & $-7.177_{\pm0.02}$ & $-9.064_{\pm0.02}$ & $-8.871_{\pm0.08}$ & $-13.816_{\pm0.40}$ \\
NSMPP$^{\mathrm{q}}$ & $-1.291_{\pm0.97}^{\dag}$ & $5.775_{\pm0.71}$ & $-7.014_{\pm1.52}^{\dag}$ & $3.898_{\pm1.08}$ & $1.954_{\pm1.62}$ & $-13.339_{\pm1.85}$ & $-7.454_{\pm0.78}$ & $-8.115_{\pm0.88}$ & $-10.266_{\pm0.90}$ & $-6.407_{\pm0.74}$ & $-7.659_{\pm0.81}$ & $-6.409_{\pm0.29}$ & $-11.703_{\pm1.05}$ \\
NJSDE$^{\mathrm{o}}$ & $-2.226_{\pm0.02}$ & $4.865_{\pm0.51}$ & $-8.960_{\pm0.01}$ & $2.646_{\pm0.26}$ & $1.473_{\pm0.02}$ & $-15.126_{\pm0.02}$ & $-8.577_{\pm0.01}$ & $-8.656_{\pm0.00}$ & $-10.913_{\pm0.14}$ & $-6.925_{\pm0.00}$ & $-8.900_{\pm0.01}^{\dag}$ & $-8.637_{\pm0.07}^{\dag}$ & $-13.616_{\pm0.00}$ \\
Attn-CNF$^{\mathrm{o}}$ & $\mathbf{-2.513_{\pm0.05}}$ & $4.397_{\pm0.09}$ & $\mathbf{-9.335_{\pm0.01}}$ & $1.773_{\pm0.04}^{n=2,\mathrm{b}\mathrm{s}}$ & $0.466_{\pm0.00}^{n=2,\mathrm{b}}$ & \underline{$-15.467_{\pm0.02}$} & $-8.773_{\pm0.02}$ & $\mathbf{-9.241_{\pm0.00}}$ & $-11.213_{\pm0.01}$ & $\mathbf{-7.349_{\pm0.02}}$ & $\mathbf{-9.264_{\pm0.05}^{\mathrm{s}\dag}}$ & $\mathbf{-9.163_{\pm0.05}^{\mathrm{s}}}$ & \underline{$-13.906_{\pm0.30}$} \\
Jump-CNF$^{\mathrm{o}}$ & \underline{$-2.496_{\pm0.01}$} & $4.169_{\pm0.07}^{\mathrm{s}}$ & \underline{$-9.326_{\pm0.01}^{\dag}$} & $1.762_{\pm0.05}^{n=2,\mathrm{d}}$ & \underline{$0.340_{\pm0.03}$} & {\scriptsize diverged} & $-8.712_{\pm0.01}$ & \underline{$-9.208_{\pm0.03}^{\mathrm{s}}$} & {\scriptsize diverged} & \underline{$-7.226_{\pm0.02}^{\dag}$} & \underline{$-9.178_{\pm0.01}^{\mathrm{s}}$} & \underline{$-8.976_{\pm0.03}$} & $\mathbf{-14.114_{\pm0.05}^{\mathrm{e}\dag}}$ \\
\midrule
\multicolumn{14}{l}{\emph{Generative}} \\
SMASH$^{\mathrm{a}}$ & \cellcolor{gray!12}$-2.365_{\pm0.08}$ & \cellcolor{gray!12}$6.293_{\pm0.59}$ & \cellcolor{gray!12}$-8.942_{\pm0.05}$ & \cellcolor{gray!12}$2.193_{\pm0.35}$ & \cellcolor{gray!12}$1.725_{\pm0.16}$ & \cellcolor{gray!12}$-15.177_{\pm0.02}$ & \cellcolor{gray!12}$-8.722_{\pm0.05}$ & \cellcolor{gray!12}$-8.907_{\pm0.06}$ & \cellcolor{gray!12}$-9.942_{\pm1.14}$ & \cellcolor{gray!12}$-6.522_{\pm0.58}$ & \cellcolor{gray!12}$-8.861_{\pm0.04}$ & \cellcolor{gray!12}$-8.409_{\pm0.13}$ & \cellcolor{gray!12}$-13.596_{\pm0.09}$ \\
DSTPP$^{\mathrm{ub}}$ & \cellcolor{gray!12}$0.874_{\pm0.54}$ & \cellcolor{gray!12}$7.970_{\pm0.48}$ & \cellcolor{gray!12}$-6.624_{\pm0.16}$ & \cellcolor{gray!12}$148.691_{\pm0.09}^{r}$ & \cellcolor{gray!12}$3.414_{\pm0.40}$ & \cellcolor{gray!12}$-10.739_{\pm0.17}$ & \cellcolor{gray!12}$-5.776_{\pm0.25}$ & \cellcolor{gray!12}$-5.949_{\pm0.28}$ & \cellcolor{gray!12}$-7.625_{\pm0.13}$ & \cellcolor{gray!12}$-4.518_{\pm0.10}$ & \cellcolor{gray!12}$-5.600_{\pm0.29}$ & \cellcolor{gray!12}$-5.136_{\pm0.25}$ & \cellcolor{gray!12}$-9.756_{\pm0.12}$ \\
\bottomrule
\end{tabular}}
\end{sidewaystable*}

\begin{figure*}[t]
\centering
\includegraphics[width=0.92\textwidth]{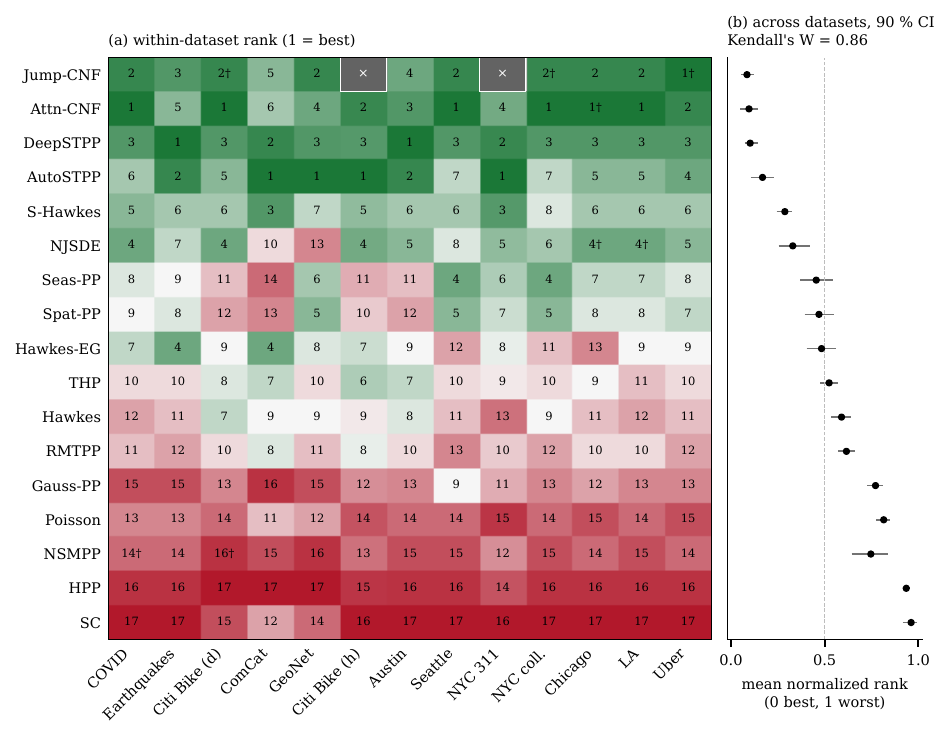}
\caption{Within-dataset NLL ranks on the real datasets: (a) rank per dataset (1 = best), (b) mean normalized rank with 90\% bootstrap interval; the ranking is fairly consistent across datasets (Kendall's $W=0.86$) with no preset best everywhere. Every reported value is ranked; \dag: numerical validation not passed; dark: diverged. Details: App.~\ref{sec:rank-notes}.}
\label{fig:rank_overview}
\end{figure*}

\begin{table}[t]
\centering
\scriptsize
\setlength{\tabcolsep}{3pt}
\caption{\rev{Chicago and Seattle: test NLL over all targets (raw) and over the targets outside the recording outages (masked; the value in Table~\ref{tab:real_data_expanded_nll}), mean $\pm$ std over 3 seeds; non-finite: every seed has non-finite targets, all inside the outages.}}
\label{tab:realdata_raw_outage}
\resizebox{\linewidth}{!}{%
\begin{tabular}{lcccc}
\toprule
& \multicolumn{2}{c}{\textbf{Chicago}} & \multicolumn{2}{c}{\textbf{Seattle}} \\
\textbf{Preset} & raw & masked & raw & masked \\
\midrule
\multicolumn{5}{l}{\emph{Classical}} \\
HPP & $-6.749_{\pm0.000}$ & $-6.764_{\pm0.000}$ & $-7.090_{\pm0.000}$ & $-7.100_{\pm0.000}$ \\
Seas-PP & $-8.370_{\pm0.000}$ & $-8.384_{\pm0.000}$ & $-8.934_{\pm0.000}$ & $-8.944_{\pm0.000}$ \\
Spat-PP & $-8.370_{\pm0.000}$ & $-8.384_{\pm0.000}$ & $-8.934_{\pm0.000}$ & $-8.943_{\pm0.000}$ \\
Hawkes-EG & $-7.830_{\pm0.000}$ & $-7.839_{\pm0.000}$ & $-8.436_{\pm0.000}$ & $-8.439_{\pm0.000}$ \\
Gauss-PP & $-7.967_{\pm0.000}$ & $-7.981_{\pm0.000}$ & $-8.435_{\pm0.000}$ & $-8.444_{\pm0.000}$ \\
\bottomrule
\end{tabular}}
\end{table}

\begin{table*}[p]
\centering
\caption{\rev{Next-event calibration from each model's compensator, without sampling: temporal PIT KS statistic (lower is better), mean $\pm$ std over 3 seeds. Bold: best per dataset; $n$: seeds if fewer than three; \dag: numerical check not passed. Further metrics: Tables~\ref{tab:calib_full_crps}--\ref{tab:calib_full_hpd_ks}; details: App.~\ref{sec:calib-notes}.}}
\label{tab:calibration_main}\label{tab:calib_full_pit_ks}
\scriptsize
\setlength{\tabcolsep}{1.5pt}
\begin{tabular}{lccccccc}
\toprule
\textbf{Preset} & \textbf{COVID} & \textbf{\shortstack{Earth-\\quakes}} & \textbf{\shortstack{Citi\\Bike (d)}} & \textbf{ComCat} & \textbf{GeoNet} & \textbf{\shortstack{Citi\\Bike (h)}} & \textbf{Austin} \\
\midrule
\multicolumn{8}{l}{\emph{Classical}} \\
HPP & ${0.317}_{\pm0.000}$ & ${0.181}_{\pm0.000}$ & ${0.187}_{\pm0.000}$ & ${0.225}_{\pm0.000}$ & ${0.033}_{\pm0.000}$ & ${0.168}_{\pm0.000}$ & ${0.077}_{\pm0.000}$ \\
Seas-PP & ${0.309}_{\pm0.000}$ & ${0.181}_{\pm0.000}$ & ${0.187}_{\pm0.000}$ & ${0.226}_{\pm0.000}$ & ${0.034}_{\pm0.000}$ & ${0.168}_{\pm0.000}$ & ${0.073}_{\pm0.000}$ \\
Spat-PP & ${0.317}_{\pm0.000}$ & ${0.181}_{\pm0.000}$ & ${0.187}_{\pm0.000}$ & ${0.225}_{\pm0.000}$ & ${0.033}_{\pm0.000}$ & ${0.168}_{\pm0.000}$ & ${0.077}_{\pm0.000}$ \\
Hawkes-EG & ${0.124}_{\pm0.000}$ & ${0.123}_{\pm0.000}$ & ${0.070}_{\pm0.000}$ & ${0.167}_{\pm0.000}$ & ${0.065}_{\pm0.000}$ & ${0.098}_{\pm0.000}$ & ${0.047}_{\pm0.000}$ \\
Gauss-PP & ${0.467}_{\pm0.016}$ & ${0.169}_{\pm0.001}$ & ${0.125}_{\pm0.001}$ & ${0.248}_{\pm0.000}$ & ${0.074}_{\pm0.017}$ & ${0.174}_{\pm0.007}$ & ${0.106}_{\pm0.008}$ \\
S-Hawkes & ${0.124}_{\pm0.004}$ & ${0.128}_{\pm0.001}$ & ${0.037}_{\pm0.002}$ & ${0.164}_{\pm0.000}$ & ${0.072}_{\pm0.002}$ & ${0.047}_{\pm0.001}$ & ${0.060}_{\pm0.002}$ \\
\bottomrule
\end{tabular}
\vspace{2pt}
\begin{tabular}{lcccccc}
\toprule
\textbf{Preset} & \textbf{Seattle} & \textbf{\shortstack{NYC\\311}} & \textbf{\shortstack{NYC\\coll.}} & \textbf{Chicago} & \textbf{LA} & \textbf{Uber} \\
\midrule
\multicolumn{7}{l}{\emph{Classical}} \\
HPP & ${0.025}_{\pm0.000}$ & ${0.141}_{\pm0.000}$ & ${0.136}_{\pm0.000}$ & ${0.314}_{\pm0.000}$ & ${0.428}_{\pm0.000}$ & ${0.222}_{\pm0.000}$ \\
Seas-PP & ${0.025}_{\pm0.000}$ & ${0.141}_{\pm0.000}$ & ${0.136}_{\pm0.000}$ & ${0.314}_{\pm0.000}$ & ${0.428}_{\pm0.000}$ & ${0.225}_{\pm0.000}$ \\
Spat-PP & ${0.025}_{\pm0.000}$ & ${0.141}_{\pm0.000}$ & ${0.136}_{\pm0.000}$ & ${0.314}_{\pm0.000}$ & ${0.428}_{\pm0.000}$ & ${0.222}_{\pm0.000}$ \\
Hawkes-EG & ${0.051}_{\pm0.000}$ & ${0.129}_{\pm0.000}$ & ${0.134}_{\pm0.000}$ & ${0.021}_{\pm0.000}$ & ${0.026}_{\pm0.000}$ & ${0.136}_{\pm0.000}$ \\
Gauss-PP & ${0.031}_{\pm0.003}$ & ${0.142}_{\pm0.001}$ & ${0.137}_{\pm0.001}$ & ${0.314}_{\pm0.001}$ & ${0.429}_{\pm0.000}$ & ${0.249}_{\pm0.007}$ \\
S-Hawkes & ${0.018}_{\pm0.003}$ & ${0.136}_{\pm0.001}$ & ${0.126}_{\pm0.001}$ & ${0.034}_{\pm0.016}$ & ${0.056}_{\pm0.007}$ & ${0.132}_{\pm0.002}$ \\
\bottomrule
\end{tabular}
\end{table*}

\begin{table*}[p]
\centering
\caption{\rev{As Table~\ref{tab:calibration_main}: CRPS$_c$ of the waiting time (hours; lower is better).}}
\label{tab:calib_full_crps}
\scriptsize
\setlength{\tabcolsep}{1.5pt}
\begin{tabular}{lccccccc}
\toprule
\textbf{Preset} & \textbf{COVID} & \textbf{\shortstack{Earth-\\quakes}} & \textbf{\shortstack{Citi\\Bike (d)}} & \textbf{ComCat} & \textbf{GeoNet} & \textbf{\shortstack{Citi\\Bike (h)}} & \textbf{Austin} \\
{\scriptsize\emph{unit}} & {\scriptsize h} & {\scriptsize h} & {\scriptsize h} & {\scriptsize h} & {\scriptsize h} & {\scriptsize $10^{-3}$\,h} & {\scriptsize h} \\
\midrule
\multicolumn{8}{l}{\emph{Classical}} \\
HPP & ${0.602}_{\pm0.000}$ & ${5.01}_{\pm0.00}$ & ${0.0886}_{\pm0.0000}$ & ${3.79}_{\pm0.00}$ & ${0.239}_{\pm0.000}$ & ${0.223}_{\pm0.000}$ & ${0.0221}_{\pm0.0000}$ \\
Seas-PP & ${0.597}_{\pm0.000}$ & ${5.02}_{\pm0.00}$ & ${0.0887}_{\pm0.0000}$ & ${3.8}_{\pm0.0}$ & ${0.239}_{\pm0.000}$ & ${0.223}_{\pm0.000}$ & ${0.0219}_{\pm0.0000}$ \\
Spat-PP & ${0.602}_{\pm0.000}$ & ${5.01}_{\pm0.00}$ & ${0.0886}_{\pm0.0000}$ & ${3.79}_{\pm0.00}$ & ${0.239}_{\pm0.000}$ & ${0.223}_{\pm0.000}$ & ${0.0221}_{\pm0.0000}$ \\
Hawkes-EG & ${0.416}_{\pm0.000}$ & ${4.48}_{\pm0.00}$ & ${0.0764}_{\pm0.0000}$ & ${3.28}_{\pm0.00}$ & ${0.238}_{\pm0.000}$ & ${2.36}_{\pm0.00}$ & ${0.019}_{\pm0.000}$ \\
Gauss-PP & ${1.01}_{\pm0.06}$ & ${4.94}_{\pm0.01}$ & ${0.0838}_{\pm0.0001}$ & ${3.76}_{\pm0.00}$ & ${0.241}_{\pm0.002}$ & ${0.224}_{\pm0.001}$ & ${0.0222}_{\pm0.0001}$ \\
\bottomrule
\end{tabular}
\vspace{2pt}
\begin{tabular}{lcccccc}
\toprule
\textbf{Preset} & \textbf{Seattle} & \textbf{\shortstack{NYC\\311}} & \textbf{\shortstack{NYC\\coll.}} & \textbf{Chicago} & \textbf{LA} & \textbf{Uber} \\
{\scriptsize\emph{unit}} & {\scriptsize h} & {\scriptsize h} & {\scriptsize h} & {\scriptsize h} & {\scriptsize h} & {\scriptsize $10^{-3}$\,h} \\
\midrule
\multicolumn{7}{l}{\emph{Classical}} \\
HPP & ${0.036}_{\pm0.000}$ & ${0.00172}_{\pm0.00000}$ & ${0.0579}_{\pm0.0000}$ & ${0.0255}_{\pm0.0000}$ & ${0.0314}_{\pm0.0000}$ & ${0.473}_{\pm0.000}$ \\
Seas-PP & ${0.036}_{\pm0.000}$ & ${0.00171}_{\pm0.00000}$ & ${0.0578}_{\pm0.0000}$ & ${0.0255}_{\pm0.0000}$ & ${0.0314}_{\pm0.0000}$ & ${0.476}_{\pm0.000}$ \\
Spat-PP & ${0.036}_{\pm0.000}$ & ${0.00172}_{\pm0.00000}$ & ${0.0579}_{\pm0.0000}$ & ${0.0255}_{\pm0.0000}$ & ${0.0314}_{\pm0.0000}$ & ${0.473}_{\pm0.000}$ \\
Hawkes-EG & ${0.0356}_{\pm0.0000}$ & ${0.00161}_{\pm0.00000}$ & ${0.0561}_{\pm0.0000}$ & ${0.022}_{\pm0.000}$ & ${0.0263}_{\pm0.0000}$ & ${0.392}_{\pm0.000}$ \\
Gauss-PP & ${0.036}_{\pm0.000}$ & ${0.00172}_{\pm0.00000}$ & ${0.0579}_{\pm0.0000}$ & ${0.0255}_{\pm0.0000}$ & ${0.0314}_{\pm0.0000}$ & ${0.495}_{\pm0.006}$ \\
\bottomrule
\end{tabular}
\end{table*}

\begin{table*}[p]
\centering
\caption{\rev{As Table~\ref{tab:calibration_main}: median absolute timing error (hours; lower is better).}}
\label{tab:calib_full_medae}
\scriptsize
\setlength{\tabcolsep}{1.5pt}
\begin{tabular}{lccccccc}
\toprule
\textbf{Preset} & \textbf{COVID} & \textbf{\shortstack{Earth-\\quakes}} & \textbf{\shortstack{Citi\\Bike (d)}} & \textbf{ComCat} & \textbf{GeoNet} & \textbf{\shortstack{Citi\\Bike (h)}} & \textbf{Austin} \\
{\scriptsize\emph{unit}} & {\scriptsize h} & {\scriptsize h} & {\scriptsize h} & {\scriptsize h} & {\scriptsize h} & {\scriptsize $10^{-3}$\,h} & {\scriptsize h} \\
\midrule
\multicolumn{8}{l}{\emph{Classical}} \\
HPP & ${0.798}_{\pm0.000}$ & ${5.42}_{\pm0.00}$ & ${0.0865}_{\pm0.0000}$ & ${2.47}_{\pm0.00}$ & ${0.219}_{\pm0.000}$ & ${0.173}_{\pm0.000}$ & ${0.013}_{\pm0.000}$ \\
Seas-PP & ${0.783}_{\pm0.000}$ & ${5.42}_{\pm0.00}$ & ${0.0863}_{\pm0.0000}$ & ${2.43}_{\pm0.00}$ & ${0.22}_{\pm0.00}$ & ${0.173}_{\pm0.000}$ & ${0.0122}_{\pm0.0000}$ \\
Spat-PP & ${0.798}_{\pm0.000}$ & ${5.42}_{\pm0.00}$ & ${0.0865}_{\pm0.0000}$ & ${2.47}_{\pm0.00}$ & ${0.219}_{\pm0.000}$ & ${0.173}_{\pm0.000}$ & ${0.013}_{\pm0.000}$ \\
Hawkes-EG & ${0.421}_{\pm0.000}$ & ${3.3}_{\pm0.0}$ & ${0.0577}_{\pm0.0000}$ & ${1.54}_{\pm0.00}$ & ${0.22}_{\pm0.00}$ & ${0.126}_{\pm0.000}$ & ${0.0102}_{\pm0.0000}$ \\
Gauss-PP & ${1.49}_{\pm0.10}$ & ${4.99}_{\pm0.03}$ & ${0.071}_{\pm0.000}$ & ${3.11}_{\pm0.00}$ & ${0.242}_{\pm0.010}$ & ${0.177}_{\pm0.004}$ & ${0.0148}_{\pm0.0004}$ \\
\bottomrule
\end{tabular}
\vspace{2pt}
\begin{tabular}{lcccccc}
\toprule
\textbf{Preset} & \textbf{Seattle} & \textbf{\shortstack{NYC\\311}} & \textbf{\shortstack{NYC\\coll.}} & \textbf{Chicago} & \textbf{LA} & \textbf{Uber} \\
{\scriptsize\emph{unit}} & {\scriptsize h} & {\scriptsize h} & {\scriptsize h} & {\scriptsize h} & {\scriptsize h} & {\scriptsize $10^{-3}$\,h} \\
\midrule
\multicolumn{7}{l}{\emph{Classical}} \\
HPP & ${0.0328}_{\pm0.0000}$ & ${0.00161}_{\pm0.00000}$ & ${0.0586}_{\pm0.0000}$ & ${0.022}_{\pm0.000}$ & ${0.0253}_{\pm0.0000}$ & ${0.513}_{\pm0.000}$ \\
Seas-PP & ${0.033}_{\pm0.000}$ & ${0.0016}_{\pm0.0000}$ & ${0.0582}_{\pm0.0000}$ & ${0.022}_{\pm0.000}$ & ${0.0253}_{\pm0.0000}$ & ${0.503}_{\pm0.000}$ \\
Spat-PP & ${0.0328}_{\pm0.0000}$ & ${0.00161}_{\pm0.00000}$ & ${0.0586}_{\pm0.0000}$ & ${0.022}_{\pm0.000}$ & ${0.0253}_{\pm0.0000}$ & ${0.513}_{\pm0.000}$ \\
Hawkes-EG & ${0.0335}_{\pm0.0000}$ & ${0.00137}_{\pm0.00000}$ & ${0.0541}_{\pm0.0000}$ & $\mathbf{{0.00843}_{\pm0.00000}}$ & ${0.00604}_{\pm0.00000}$ & ${0.368}_{\pm0.000}$ \\
Gauss-PP & ${0.0333}_{\pm0.0002}$ & ${0.00161}_{\pm0.00001}$ & ${0.059}_{\pm0.000}$ & ${0.022}_{\pm0.000}$ & ${0.0255}_{\pm0.0000}$ & ${0.562}_{\pm0.014}$ \\
\bottomrule
\end{tabular}
\end{table*}

\begin{table*}[p]
\centering
\caption{\rev{As Table~\ref{tab:calibration_main}: coverage of the central 80\% interval (closest to 0.80 is best).}}
\label{tab:calib_full_cov80}
\scriptsize
\setlength{\tabcolsep}{1.5pt}
\begin{tabular}{lccccccc}
\toprule
\textbf{Preset} & \textbf{COVID} & \textbf{\shortstack{Earth-\\quakes}} & \textbf{\shortstack{Citi\\Bike (d)}} & \textbf{ComCat} & \textbf{GeoNet} & \textbf{\shortstack{Citi\\Bike (h)}} & \textbf{Austin} \\
\midrule
\multicolumn{8}{l}{\emph{Classical}} \\
HPP & ${0.754}_{\pm0.000}$ & ${0.648}_{\pm0.000}$ & ${0.768}_{\pm0.000}$ & ${0.465}_{\pm0.000}$ & ${0.788}_{\pm0.000}$ & ${0.765}_{\pm0.000}$ & ${0.685}_{\pm0.000}$ \\
Seas-PP & ${0.756}_{\pm0.000}$ & ${0.648}_{\pm0.000}$ & ${0.767}_{\pm0.000}$ & ${0.464}_{\pm0.000}$ & ${0.787}_{\pm0.000}$ & ${0.765}_{\pm0.000}$ & ${0.690}_{\pm0.000}$ \\
Spat-PP & ${0.754}_{\pm0.000}$ & ${0.648}_{\pm0.000}$ & ${0.768}_{\pm0.000}$ & ${0.465}_{\pm0.000}$ & ${0.788}_{\pm0.000}$ & ${0.765}_{\pm0.000}$ & ${0.685}_{\pm0.000}$ \\
Hawkes-EG & ${0.824}_{\pm0.000}$ & ${0.681}_{\pm0.000}$ & ${0.813}_{\pm0.000}$ & ${0.556}_{\pm0.000}$ & ${0.827}_{\pm0.000}$ & ${0.825}_{\pm0.000}$ & ${0.738}_{\pm0.000}$ \\
Gauss-PP & ${0.645}_{\pm0.015}$ & ${0.637}_{\pm0.001}$ & ${0.773}_{\pm0.000}$ & ${0.490}_{\pm0.000}$ & ${0.797}_{\pm0.004}$ & ${0.764}_{\pm0.001}$ & ${0.694}_{\pm0.002}$ \\
S-Hawkes & ${0.826}_{\pm0.000}$ & ${0.686}_{\pm0.001}$ & ${0.798}_{\pm0.001}$ & ${0.557}_{\pm0.001}$ & ${0.825}_{\pm0.000}$ & ${0.807}_{\pm0.001}$ & ${0.718}_{\pm0.001}$ \\
\bottomrule
\end{tabular}
\vspace{2pt}
\begin{tabular}{lcccccc}
\toprule
\textbf{Preset} & \textbf{Seattle} & \textbf{\shortstack{NYC\\311}} & \textbf{\shortstack{NYC\\coll.}} & \textbf{Chicago} & \textbf{LA} & \textbf{Uber} \\
\midrule
\multicolumn{7}{l}{\emph{Classical}} \\
HPP & $\mathbf{{0.801}_{\pm0.000}}$ & ${0.658}_{\pm0.000}$ & ${0.655}_{\pm0.000}$ & ${0.440}_{\pm0.000}$ & ${0.344}_{\pm0.000}$ & ${0.769}_{\pm0.000}$ \\
Seas-PP & ${0.802}_{\pm0.000}$ & ${0.659}_{\pm0.000}$ & ${0.656}_{\pm0.000}$ & ${0.440}_{\pm0.000}$ & ${0.342}_{\pm0.000}$ & ${0.770}_{\pm0.000}$ \\
Spat-PP & $\mathbf{{0.801}_{\pm0.000}}$ & ${0.658}_{\pm0.000}$ & ${0.655}_{\pm0.000}$ & ${0.440}_{\pm0.000}$ & ${0.344}_{\pm0.000}$ & ${0.769}_{\pm0.000}$ \\
Hawkes-EG & ${0.828}_{\pm0.000}$ & ${0.687}_{\pm0.000}$ & ${0.698}_{\pm0.000}$ & $\mathbf{{0.794}_{\pm0.000}}$ & ${0.785}_{\pm0.000}$ & ${0.818}_{\pm0.000}$ \\
Gauss-PP & ${0.804}_{\pm0.001}$ & ${0.658}_{\pm0.000}$ & ${0.656}_{\pm0.001}$ & ${0.440}_{\pm0.001}$ & ${0.345}_{\pm0.000}$ & ${0.761}_{\pm0.002}$ \\
S-Hawkes & ${0.808}_{\pm0.001}$ & ${0.665}_{\pm0.001}$ & ${0.674}_{\pm0.002}$ & ${0.765}_{\pm0.014}$ & ${0.767}_{\pm0.006}$ & ${0.816}_{\pm0.001}$ \\
\bottomrule
\end{tabular}
\end{table*}

\begin{table*}[p]
\centering
\caption{\rev{As Table~\ref{tab:calibration_main}: spatial HPD-PIT KS statistic (lower is better).}}
\label{tab:calib_full_hpd_ks}
\scriptsize
\setlength{\tabcolsep}{1.5pt}
\begin{tabular}{lccccccc}
\toprule
\textbf{Preset} & \textbf{COVID} & \textbf{\shortstack{Earth-\\quakes}} & \textbf{\shortstack{Citi\\Bike (d)}} & \textbf{ComCat} & \textbf{GeoNet} & \textbf{\shortstack{Citi\\Bike (h)}} & \textbf{Austin} \\
\midrule
\multicolumn{8}{l}{\emph{Classical}} \\
HPP & {\scriptsize n/i} & {\scriptsize n/i} & {\scriptsize n/i} & {\scriptsize n/i} & {\scriptsize n/i} & {\scriptsize n/i} & {\scriptsize n/i} \\
Seas-PP & ${0.105}_{\pm0.000}$ & ${0.064}_{\pm0.000}$ & $\mathbf{{0.007}_{\pm0.000}}$ & ${0.168}_{\pm0.000}$ & ${0.059}_{\pm0.000}$ & $\mathbf{{0.005}_{\pm0.000}}$ & ${0.053}_{\pm0.000}$ \\
Spat-PP & ${0.105}_{\pm0.000}$ & ${0.064}_{\pm0.000}$ & $\mathbf{{0.007}_{\pm0.000}}$ & ${0.168}_{\pm0.000}$ & ${0.059}_{\pm0.000}$ & $\mathbf{{0.005}_{\pm0.000}}$ & ${0.053}_{\pm0.000}$ \\
Hawkes-EG & ${0.157}_{\pm0.000}$ & ${0.175}_{\pm0.000}$ & ${0.094}_{\pm0.000}$ & $\mathbf{{0.057}_{\pm0.000}}$ & ${0.090}_{\pm0.000}$ & ${0.094}_{\pm0.000}$ & ${0.047}_{\pm0.000}$ \\
Gauss-PP & ${0.171}_{\pm0.007}$ & ${0.155}_{\pm0.003}$ & ${0.056}_{\pm0.005}$ & ${0.226}_{\pm0.008}$ & ${0.265}_{\pm0.005}$ & ${0.055}_{\pm0.005}$ & ${0.100}_{\pm0.005}$ \\
S-Hawkes & ${0.153}_{\pm0.012}$ & ${0.183}_{\pm0.005}$ & ${0.048}_{\pm0.001}$ & ${0.082}_{\pm0.003}$ & ${0.089}_{\pm0.000}$ & ${0.033}_{\pm0.003}$ & ${0.054}_{\pm0.001}$ \\
\bottomrule
\end{tabular}
\vspace{2pt}
\begin{tabular}{lcccccc}
\toprule
\textbf{Preset} & \textbf{Seattle} & \textbf{\shortstack{NYC\\311}} & \textbf{\shortstack{NYC\\coll.}} & \textbf{Chicago} & \textbf{LA} & \textbf{Uber} \\
\midrule
\multicolumn{7}{l}{\emph{Classical}} \\
HPP & {\scriptsize n/i} & {\scriptsize n/i} & {\scriptsize n/i} & {\scriptsize n/i} & {\scriptsize n/i} & {\scriptsize n/i} \\
Seas-PP & $\mathbf{{0.010}_{\pm0.000}}$ & $\mathbf{{0.035}_{\pm0.000}}$ & $\mathbf{{0.021}_{\pm0.000}}$ & $\mathbf{{0.008}_{\pm0.000}}$ & $\mathbf{{0.006}_{\pm0.000}}$ & ${0.047}_{\pm0.000}$ \\
Spat-PP & $\mathbf{{0.010}_{\pm0.000}}$ & $\mathbf{{0.035}_{\pm0.000}}$ & $\mathbf{{0.021}_{\pm0.000}}$ & $\mathbf{{0.008}_{\pm0.000}}$ & $\mathbf{{0.006}_{\pm0.000}}$ & ${0.047}_{\pm0.000}$ \\
Hawkes-EG & ${0.135}_{\pm0.000}$ & ${0.052}_{\pm0.000}$ & ${0.024}_{\pm0.000}$ & ${0.162}_{\pm0.000}$ & ${0.114}_{\pm0.000}$ & ${0.098}_{\pm0.000}$ \\
Gauss-PP & ${0.179}_{\pm0.004}$ & ${0.064}_{\pm0.004}$ & ${0.036}_{\pm0.000}$ & ${0.070}_{\pm0.000}$ & ${0.066}_{\pm0.001}$ & ${0.289}_{\pm0.003}$ \\
S-Hawkes & ${0.092}_{\pm0.003}$ & ${0.098}_{\pm0.009}$ & ${0.036}_{\pm0.003}$ & ${0.048}_{\pm0.003}$ & ${0.060}_{\pm0.001}$ & ${0.050}_{\pm0.001}$ \\
\bottomrule
\end{tabular}
\end{table*}

\begin{table*}[t]
\centering
\footnotesize
\setlength{\tabcolsep}{4pt}
\caption{\rev{Next-event forecasts from $K=128$ samples per target, for every preset including SMASH and DSTPP: energy score ES of the time and location (lower is better) and hit rate (higher is better; not on ComCat and GeoNet), mean over seeds. Bold and underline: best and second best per column; $n$: seeds if fewer than three; \dag: numerical check not passed, as in the NLL tables. Details: App.~\ref{sec:nextevent-notes}.}}
\label{tab:nextevent_main}\label{tab:nextevent_full}
\begin{tabular}{lcccccccc}
\toprule
& \multicolumn{1}{c}{\textbf{ComCat}} & \multicolumn{1}{c}{\textbf{GeoNet}} & \multicolumn{2}{c}{\textbf{Citi Bike (h)}} & \multicolumn{2}{c}{\textbf{Austin}} & \multicolumn{2}{c}{\textbf{Seattle}} \\
\cmidrule(lr){2-2}\cmidrule(lr){3-3}\cmidrule(lr){4-5}\cmidrule(lr){6-7}\cmidrule(lr){8-9}
\textbf{Preset} & ES & ES & ES & Hit & ES & Hit & ES & Hit \\
\midrule
\multicolumn{9}{l}{\emph{Classical}} \\
HPP & ${331}$ & ${406}$ & ${7.39}$ & ${0.024}$ & ${10.2}$ & ${0.004}$ & ${5.18}$ & ${0.026}$ \\
Seas-PP & ${262}$ & ${195}$ & ${2.97}$ & ${0.056}$ & ${7.48}$ & ${0.009}$ & \underline{${4.09}$} & ${0.084}$ \\
Spat-PP & ${262}$ & ${195}$ & ${2.97}$ & ${0.057}$ & ${7.49}$ & ${0.008}$ & \underline{${4.09}$} & ${0.085}$ \\
Hawkes-EG & ${198}$ & ${199}$ & ${13.1}$ & ${0.067}$ & ${7.53}$ & \underline{${0.011}$} & ${4.24}$ & ${0.056}$ \\
Gauss-PP & ${277}$ & ${203}$ & ${2.98}$ & ${0.049}$ & ${7.51}$ & ${0.007}$ & ${4.12}$ & ${0.054}$ \\
\bottomrule
\end{tabular}
\vspace{4pt}
\begin{tabular}{lcccccccccc}
\toprule
& \multicolumn{2}{c}{\textbf{NYC 311}} & \multicolumn{2}{c}{\textbf{NYC coll.}} & \multicolumn{2}{c}{\textbf{Chicago}} & \multicolumn{2}{c}{\textbf{LA}} & \multicolumn{2}{c}{\textbf{Uber}} \\
\cmidrule(lr){2-3}\cmidrule(lr){4-5}\cmidrule(lr){6-7}\cmidrule(lr){8-9}\cmidrule(lr){10-11}
\textbf{Preset} & ES & Hit & ES & Hit & ES & Hit & ES & Hit & ES & Hit \\
\midrule
\multicolumn{11}{l}{\emph{Classical}} \\
HPP & ${9.54}$ & ${0.004}$ & ${9.53}$ & \underline{${0.004}$} & ${9.99}$ & ${0.000}$ & ${17.4}$ & \underline{${0.001}$} & ${7.55}$ & ${0.029}$ \\
Seas-PP & ${7.26}$ & ${0.001}$ & ${7.40}$ & ${0.003}$ & ${8.61}$ & ${0.000}$ & ${14.8}$ & ${0.000}$ & ${3.59}$ & ${0.058}$ \\
Spat-PP & ${7.26}$ & ${0.001}$ & ${7.40}$ & \underline{${0.004}$} & ${8.61}$ & ${0.000}$ & ${14.8}$ & ${0.000}$ & ${3.59}$ & ${0.053}$ \\
Hawkes-EG & ${7.49}$ & \underline{${0.005}$} & ${7.65}$ & \underline{${0.004}$} & ${9.30}$ & $\mathbf{{0.004}}$ & ${15.5}$ & \underline{${0.001}$} & ${3.61}$ & ${0.074}$ \\
Gauss-PP & ${7.33}$ & ${0.001}$ & ${7.49}$ & ${0.002}$ & ${8.68}$ & ${0.000}$ & ${14.9}$ & ${0.000}$ & ${3.75}$ & ${0.016}$ \\
\bottomrule
\end{tabular}
\end{table*}

\clearpage
\begin{table}[t]
\centering
\scriptsize
\caption{\rev{Mean rank over the ten benchmark datasets: energy score, hit rate (8 datasets) and, for the presets with a likelihood, NLL. Superscript: number of datasets when fewer. Spearman correlation of the ES and NLL rankings: median $0.88$ (range $0.70$--$0.95$).}}
\label{tab:nextevent_ranks}
\begin{tabular}{lccc@{\hspace{14pt}}lccc}
\toprule
\textbf{Preset} & \textbf{ES} & \textbf{Hit} & \textbf{NLL} & \textbf{Preset} & \textbf{ES} & \textbf{Hit} & \textbf{NLL} \\
\midrule
HPP & 17.5 & 11.4 & 15.8 & THP & 8.9 & 8.9 & 8.9 \\
Seas-PP & 7.5 & 10.6 & 7.6 & AutoSTPP & 8.3 & 6.0 & 3.2 \\
Spat-PP & 7.9 & 10.4 & 7.8 & DeepSTPP & 2.1 & 4.4 & 5.0 \\
Hawkes-EG & 13.3 & 6.0 & 8.6 & NSMPP & 16.3 & 13.1 & 14.3 \\
Gauss-PP & 11.4 & 13.8 & 12.7 & NJSDE & 5.3 & 6.0 & 6.2 \\
S-Hawkes & 8.0 & 6.2 & 5.4 & Attn-CNF & 2.5 & 6.1 & 2.5 \\
Poisson & 15.4 & 14.8 & 13.6 & Jump-CNF & 2.6$^{(8)}$ & 7.3$^{(6)}$ & 2.5 \\
Hawkes & 11.0 & 10.5 & 9.8 & SMASH & 7.2 & 3.8 & --- \\
SC & 17.1 & 16.6 & 16.0 & DSTPP & 11.7 & 12.9 & --- \\
RMTPP & 12.7 & 9.1 & 10.2 & & & & \\
\bottomrule
\end{tabular}
\end{table}

\subsection{\rev{HawkesNest: Temporal Calibration}}
\label{app:additional-learning-curves}

\rev{Table~\ref{tab:suite3_v2_pit} reports the temporal calibration of the next-event waiting time on the entanglement suite. The PIT KS statistic and CRPS$_c$ show no trend across levels beyond seed variation (seed standard deviation of the PIT KS statistic up to 0.08 for NSMPP and 0.10 for DSTPP); AutoSTPP and NJSDE are the two best at every level, and RMTPP and THP stay poorly calibrated (PIT KS about 0.5). One-step timing calibration therefore does not register the coupling, which appears in the likelihood and in intensity recovery (Section~\ref{sec:experiments}).}

\begin{table}[t]
\centering
\small
\setlength{\tabcolsep}{3pt}
\caption{\rev{HawkesNest: temporal calibration of the waiting time on the common targets, PIT KS statistic and CRPS$_c$ ($c=2$), mean over 3 seeds, lower is better. Bold and underline: best and second-best value per level among the likelihood models (ties at the printed precision share the mark). Generative models: estimated from 200 samples per target, not ranked.}}
\label{tab:suite3_v2_pit}
\begin{tabular}{lcccc@{\hspace{10pt}}cccc}
\toprule
& \multicolumn{4}{c}{PIT KS} & \multicolumn{4}{c}{CRPS$_c$} \\
\cmidrule(lr){2-5}\cmidrule(lr){6-9}
Model & L0 & L1 & L2 & L3 & L0 & L1 & L2 & L3 \\
\midrule
\multicolumn{9}{l}{\emph{Neural STPP}} \\
AutoSTPP & $\mathbf{0.017}$ & $\mathbf{0.012}$ & $\mathbf{0.017}$ & $\underline{0.015}$ & $\mathbf{0.0360}$ & $\mathbf{0.0354}$ & $\mathbf{0.0354}$ & $\mathbf{0.0352}$ \\
DeepSTPP & $0.071$ & $0.073$ & $\underline{0.068}$ & $0.075$ & $0.0377$ & $0.0370$ & $\underline{0.0372}$ & $0.0373$ \\
NSMPP & $0.091$ & $0.063$ & $0.089$ & $0.109$ & $0.0391$ & $0.0378$ & $0.0395$ & $0.0393$ \\
NJSDE & $\underline{0.020}$ & $\underline{0.020}$ & $\mathbf{0.017}$ & $\mathbf{0.014}$ & $\underline{0.0361}$ & $\underline{0.0355}$ & $\mathbf{0.0354}$ & $\underline{0.0353}$ \\
\midrule
\multicolumn{9}{l}{\emph{Factorized}} \\
RMTPP & $0.523$ & $0.537$ & $0.537$ & $0.545$ & $0.1161$ & $0.1181$ & $0.1172$ & $0.1189$ \\
THP & $0.481$ & $0.493$ & $0.494$ & $0.505$ & $0.0960$ & $0.0976$ & $0.0970$ & $0.0985$ \\
\midrule
\multicolumn{9}{l}{\emph{Generative (sample-based, not ranked)}} \\
SMASH & \cellcolor{gray!12}$0.028$ & \cellcolor{gray!12}$0.034$ & \cellcolor{gray!12}$0.023$ & \cellcolor{gray!12}$0.034$ & \cellcolor{gray!12}$0.0379$ & \cellcolor{gray!12}$0.0372$ & \cellcolor{gray!12}$0.0376$ & \cellcolor{gray!12}$0.0376$ \\
DSTPP & \cellcolor{gray!12}$0.180$ & \cellcolor{gray!12}$0.180$ & \cellcolor{gray!12}$0.122$ & \cellcolor{gray!12}$0.144$ & \cellcolor{gray!12}$0.0389$ & \cellcolor{gray!12}$0.0385$ & \cellcolor{gray!12}$0.0379$ & \cellcolor{gray!12}$0.0381$ \\
\bottomrule
\end{tabular}
\end{table}

\clearpage
\subsection{\rev{Tuning, Training Budget, and Compute}}
\label{app:training-time-diagnostics}

\begin{figure}[t]
\centering
\begin{subfigure}[t]{0.78\linewidth}\centering\includegraphics[width=\linewidth]{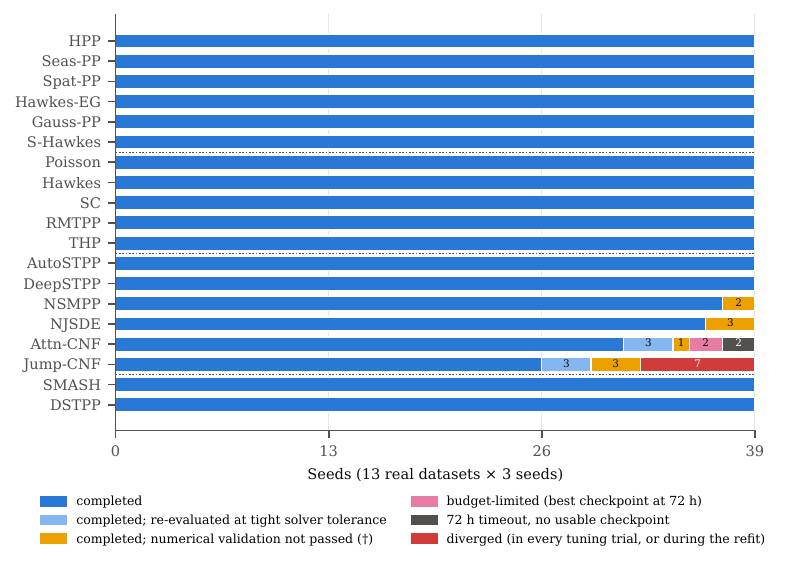}\caption{\rev{Final state of every seed}}\label{fig:runtime-reliability}\end{subfigure}

\begin{subfigure}[t]{\linewidth}\centering\includegraphics[width=\linewidth]{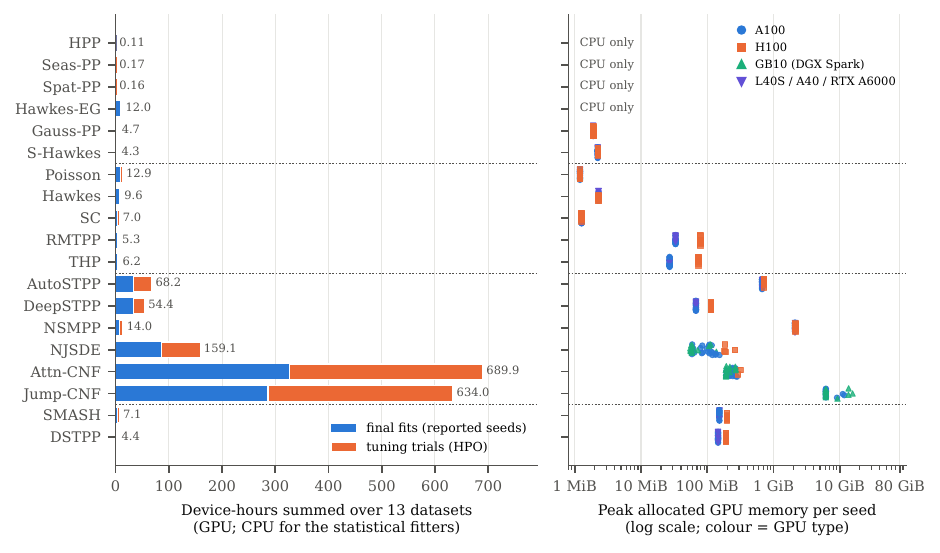}\caption{\rev{Compute and peak GPU memory}}\label{fig:runtime-compute}\end{subfigure}
\caption{\rev{Real-data runs per preset. (a) Final state of the 39 seeds (13 datasets $\times$ 3); each seed has one state. (b) Device-hours of the reported fits and tuning trials, without retries (1{,}694\,h in total), and peak GPU memory per seed, coloured by GPU type.}}
\label{fig:runtime}
\end{figure}

\rev{Table~\ref{tab:hpo_budget} gives the tuning budget. Each gradient-trained preset runs three trials of at most 5{,}000 optimization steps that scale its base learning rate, selected by validation NLL, and then refits with seeds 11, 22 and 33 for at most 20{,}000 steps and 100 epochs (batch size 64, validation every 32 steps, early stopping after 16 validations without improvement, validation-best checkpoint, 72\,h limit on the cluster). Table~\ref{tab:preset_audit} is the per-preset audit; in its seed columns, \emph{caveat} means re-evaluated with a tightened solver, numerical check not passed, or stopped at 72\,h, and \emph{excl.}\ means excluded after the 72\,h limit without a usable checkpoint; times are wall-clock on heterogeneous GPUs. Figure~\ref{fig:runtime} shows the state of every seed (a) and the compute (b): the reported fits and tuning trials took about 1{,}694 device-hours, 88\% of it for NJSDE, Attn-CNF and Jump-CNF. Table~\ref{tab:realdata_budget} shows how the training runs ended: 450 of 499 stopped early. The factorized Poisson preset often exhausts its step budget and is still improving on 9 of the 13 datasets, gaining a median 0.095 nats/event over the last 10\% of the budget, so its NLL may be budget-limited on those datasets; for the factorized Hawkes preset (5 datasets) the median gain is 0.004 nats/event.}

\begin{table}[t]
\centering
\footnotesize
\setlength{\tabcolsep}{3pt}
\caption{\rev{Tuning budget of every preset on the real datasets, identical for every dataset; the protocol is in Appendix~\ref{app:training-time-diagnostics}. Statistical fitters have no step budget; their trials are listed.}}
\label{tab:hpo_budget}
\begin{tabularx}{\linewidth}{l X c c l}
\toprule
\rev{Preset} & \rev{Tuned hyperparameter / search space} & \rev{Trials} & \rev{Steps/trial} & \rev{Optimizer, base lr} \\
\midrule
\multicolumn{5}{l}{\emph{Classical}} \\
\leavevmode\rev{HPP} & \rev{none (closed-form MLE)} & \rev{1} & \rev{---} & \rev{---} \\
\leavevmode\rev{Seas-PP} & \rev{spatial grid $\in\{8{\times}8$, $4{\times}4$, $16{\times}16\}$} & \rev{3} & \rev{---} & \rev{---} \\
\leavevmode\rev{Spat-PP} & \rev{spatial grid $\in\{8{\times}8$, $4{\times}4$, $16{\times}16\}$} & \rev{3} & \rev{---} & \rev{---} \\
\leavevmode\rev{Hawkes-EG} & \rev{none (Nelder--Mead MLE, $\le$1{,}200 iterations)} & \rev{1} & \rev{---} & \rev{---} \\
\leavevmode\rev{Gauss-PP} & \rev{learning rate $\in\{1, 0.3, 3\}\times$ base} & \rev{3} & \rev{5{,}000} & \rev{AdamW, $10^{-3}$} \\
\leavevmode\rev{S-Hawkes} & \rev{learning rate $\in\{1, 0.3, 3\}\times$ base} & \rev{3} & \rev{5{,}000} & \rev{AdamW, $10^{-3}$} \\
\midrule
\multicolumn{5}{l}{\emph{Factorized}} \\
\leavevmode\rev{Poisson} & \rev{learning rate $\in\{1, 0.3, 3\}\times$ base} & \rev{3} & \rev{5{,}000} & \rev{AdamW, $10^{-3}$} \\
\leavevmode\rev{Hawkes} & \rev{learning rate $\in\{1, 0.3, 3\}\times$ base} & \rev{3} & \rev{5{,}000} & \rev{AdamW, $10^{-3}$} \\
\leavevmode\rev{SC} & \rev{learning rate $\in\{1, 0.3, 3\}\times$ base} & \rev{3} & \rev{5{,}000} & \rev{AdamW, $10^{-3}$} \\
\leavevmode\rev{RMTPP} & \rev{learning rate $\in\{1, 0.3, 3\}\times$ base} & \rev{3} & \rev{5{,}000} & \rev{Adam, $10^{-3}$} \\
\leavevmode\rev{THP} & \rev{learning rate $\in\{1, 0.3, 3\}\times$ base} & \rev{3} & \rev{5{,}000} & \rev{Adam, $10^{-3}$} \\
\midrule
\multicolumn{5}{l}{\emph{Neural STPP}} \\
\leavevmode\rev{AutoSTPP} & \rev{learning rate $\in\{1, 0.3, 3\}\times$ base} & \rev{3} & \rev{5{,}000} & \rev{Adam, $4\cdot10^{-3}$} \\
\leavevmode\rev{DeepSTPP} & \rev{learning rate $\in\{1, 0.3, 3\}\times$ base} & \rev{3} & \rev{5{,}000} & \rev{Adam, $3\cdot10^{-4}$} \\
\leavevmode\rev{NSMPP} & \rev{learning rate $\in\{1, 0.3, 3\}\times$ base} & \rev{3} & \rev{5{,}000} & \rev{Adadelta, 1} \\
\leavevmode\rev{NJSDE} & \rev{learning rate $\in\{1, 0.3, 3\}\times$ base} & \rev{3} & \rev{5{,}000} & \rev{AdamW, $10^{-3}$} \\
\leavevmode\rev{Attn-CNF} & \rev{learning rate $\in\{1, 0.3, 3\}\times$ base} & \rev{3} & \rev{5{,}000} & \rev{AdamW, $10^{-3}$} \\
\leavevmode\rev{Jump-CNF} & \rev{learning rate $\in\{1, 0.3, 3\}\times$ base} & \rev{3} & \rev{5{,}000} & \rev{AdamW, $5\cdot10^{-4}$} \\
\midrule
\multicolumn{5}{l}{\emph{Generative}} \\
\leavevmode\rev{SMASH} & \rev{learning rate $\in\{1, 0.3, 3\}\times$ base} & \rev{3} & \rev{5{,}000} & \rev{AdamW, $10^{-3}$} \\
\leavevmode\rev{DSTPP} & \rev{learning rate $\in\{1, 0.3, 3\}\times$ base} & \rev{3} & \rev{5{,}000} & \rev{Adam, $10^{-3}$} \\
\bottomrule
\end{tabularx}
\end{table}

\begin{table}[t]
\centering
\footnotesize
\setlength{\tabcolsep}{3pt}
\caption{\rev{Per-preset audit on the 13 real datasets: likelihood type, tuning, final state of the 39 seeds, median training time per seed and largest peak GPU memory of a seed. Column meanings: Appendix~\ref{app:training-time-diagnostics}.}}
\label{tab:preset_audit}
\begin{tabular}{lllcccccc}
\toprule
& & & \multicolumn{4}{c}{\rev{Seeds (of 39)}} & & \\
\cmidrule(lr){4-7}
\rev{Preset} & \rev{Likelihood} & \rev{Tuning} & \rev{done} & \rev{caveat} & \rev{excl.} & \rev{div.} & \rev{Time/seed} & \rev{GPU mem.} \\
\midrule
\multicolumn{9}{l}{\emph{Classical}} \\
\leavevmode\rev{HPP} & \rev{exact} & \rev{fixed} & \rev{39} & \rev{0} & \rev{0} & \rev{0} & \rev{4\,s} & \rev{CPU} \\
\leavevmode\rev{Seas-PP} & \rev{exact} & \rev{grid, 3 trials} & \rev{39} & \rev{0} & \rev{0} & \rev{0} & \rev{4\,s} & \rev{CPU} \\
\leavevmode\rev{Spat-PP} & \rev{exact} & \rev{grid, 3 trials} & \rev{39} & \rev{0} & \rev{0} & \rev{0} & \rev{4\,s} & \rev{CPU} \\
\leavevmode\rev{Hawkes-EG} & \rev{exact} & \rev{fixed} & \rev{39} & \rev{0} & \rev{0} & \rev{0} & \rev{1\,min} & \rev{CPU} \\
\leavevmode\rev{Gauss-PP} & \rev{exact} & \rev{lr, 3 trials} & \rev{39} & \rev{0} & \rev{0} & \rev{0} & \rev{3\,min} & \rev{2\,MiB} \\
\leavevmode\rev{S-Hawkes} & \rev{exact} & \rev{lr, 3 trials} & \rev{39} & \rev{0} & \rev{0} & \rev{0} & \rev{3\,min} & \rev{2\,MiB} \\
\midrule
\multicolumn{9}{l}{\emph{Factorized}} \\
\leavevmode\rev{Poisson} & \rev{exact} & \rev{lr, 3 trials} & \rev{39} & \rev{0} & \rev{0} & \rev{0} & \rev{9\,min} & \rev{1\,MiB} \\
\leavevmode\rev{Hawkes} & \rev{exact} & \rev{lr, 3 trials} & \rev{39} & \rev{0} & \rev{0} & \rev{0} & \rev{7\,min} & \rev{2\,MiB} \\
\leavevmode\rev{SC} & \rev{exact} & \rev{lr, 3 trials} & \rev{39} & \rev{0} & \rev{0} & \rev{0} & \rev{5\,min} & \rev{1\,MiB} \\
\leavevmode\rev{RMTPP} & \rev{exact} & \rev{lr, 3 trials} & \rev{39} & \rev{0} & \rev{0} & \rev{0} & \rev{4\,min} & \rev{80\,MiB} \\
\leavevmode\rev{THP} & \rev{quadrature} & \rev{lr, 3 trials} & \rev{39} & \rev{0} & \rev{0} & \rev{0} & \rev{3\,min} & \rev{75\,MiB} \\
\midrule
\multicolumn{9}{l}{\emph{Neural STPP}} \\
\leavevmode\rev{AutoSTPP} & \rev{exact} & \rev{lr, 3 trials} & \rev{39} & \rev{0} & \rev{0} & \rev{0} & \rev{26\,min} & \rev{736\,MiB} \\
\leavevmode\rev{DeepSTPP} & \rev{exact} & \rev{lr, 3 trials} & \rev{39} & \rev{0} & \rev{0} & \rev{0} & \rev{7\,min} & \rev{116\,MiB} \\
\leavevmode\rev{NSMPP} & \rev{quadrature} & \rev{lr, 3 trials} & \rev{37} & \rev{2} & \rev{0} & \rev{0} & \rev{11\,min} & \rev{2.2\,GiB} \\
\leavevmode\rev{NJSDE} & \rev{ODE solver} & \rev{lr, 3 trials} & \rev{36} & \rev{3} & \rev{0} & \rev{0} & \rev{1.7\,h} & \rev{269\,MiB} \\
\leavevmode\rev{Attn-CNF} & \rev{ODE solver} & \rev{lr, 3 trials} & \rev{31} & \rev{6} & \rev{2} & \rev{0} & \rev{6.1\,h} & \rev{326\,MiB} \\
\leavevmode\rev{Jump-CNF} & \rev{ODE solver} & \rev{lr, 3 trials} & \rev{26} & \rev{6} & \rev{0} & \rev{7} & \rev{6.9\,h} & \rev{15.6\,GiB} \\
\midrule
\multicolumn{9}{l}{\emph{Generative}} \\
\leavevmode\rev{SMASH} & \rev{approximation} & \rev{lr, 3 trials} & \rev{39} & \rev{0} & \rev{0} & \rev{0} & \rev{5\,min} & \rev{203\,MiB} \\
\leavevmode\rev{DSTPP} & \rev{bound} & \rev{lr, 3 trials} & \rev{39} & \rev{0} & \rev{0} & \rev{0} & \rev{3\,min} & \rev{197\,MiB} \\
\bottomrule
\end{tabular}
\end{table}

\begin{table}[t]
\centering
\small
\setlength{\tabcolsep}{4pt}
\caption{\rev{How the real-data training runs ended, per preset (13 datasets $\times$ 3 seeds; runs without training logs omitted). 72\,h: stopped by the wall-clock limit; diverged: during the refit. Still improving: datasets on which most seeds have their best per-event validation NLL in the last 10\% of the budget, within the patience window before an early stop, or in the last 10\% of the executed steps of a run stopped at 72\,h. NSMPP selects its checkpoint on a per-sequence criterion while its per-event validation NLL is still decreasing; the selected checkpoint is a median 0.067 nats/event worse than its best per-event one.}}
\label{tab:realdata_budget}
\begin{tabular}{lcccccc}
\toprule
\rev{Preset} & \rev{runs} & \rev{early stop} & \rev{budget} & \rev{72\,h} & \rev{diverged} & \rev{still improving} \\
\midrule
\multicolumn{7}{l}{\emph{Classical (gradient-trained)}} \\
Gauss-PP & 39 & 39 & 0 & 0 & 0 & 0 / 13 \\
S-Hawkes & 39 & 39 & 0 & 0 & 0 & 0 / 13 \\
\midrule
\multicolumn{7}{l}{\emph{Factorized}} \\
Poisson & 39 & 12 & 27 & 0 & 0 & 9 / 13 \\
Hawkes & 39 & 26 & 13 & 0 & 0 & 5 / 13 \\
SC & 39 & 36 & 3 & 0 & 0 & 1 / 13 \\
RMTPP & 39 & 36 & 3 & 0 & 0 & 1 / 13 \\
THP & 39 & 39 & 0 & 0 & 0 & 0 / 13 \\
\midrule
\multicolumn{7}{l}{\emph{Neural STPP}} \\
AutoSTPP & 39 & 39 & 0 & 0 & 0 & 0 / 13 \\
DeepSTPP & 39 & 39 & 0 & 0 & 0 & 0 / 13 \\
NSMPP & 39 & 39 & 0 & 0 & 0 & 9 / 13 \\
NJSDE & 39 & 39 & 0 & 0 & 0 & 0 / 13 \\
Attn-CNF & 37 & 35 & 0 & 2 & 0 & 1 / 13 \\
Jump-CNF & 33 & 32 & 0 & 0 & 1 & 0 / 11 \\
\bottomrule
\end{tabular}
\end{table}

\rev{\clearpage}
\subsection{Table and figure notes}
\label{sec:table-notes}

\paragraph{NLL tables (Tables~\ref{tab:real_data_expanded_nll}, \ref{tab:real_data_full_nll}, \ref{tab:realdata_raw_outage}).}
\rev{Test next-event NLL per event in raw units (time in days, space in degrees); every model is scored on the same targets, each conditioned on its 20 preceding events. COVID, Earthquakes and Citi Bike (d) keep the original splits of the Neural STPP benchmark, whose windows overlap in time (6.7\% of the Earthquakes test events also lie inside training windows; the COVID validation and test windows lie inside the training period), so their values are not comparable with those of the other ten datasets. Bold and underline mark the best and second-best value per dataset among the presets with an exact or numerical likelihood and at least two seeds. \emph{Markers in Table~\ref{tab:real_data_full_nll}.} After the preset name: q = compensator by numerical quadrature, o = likelihood integrated by an ODE solver. In a cell: $n$ = number of seeds if fewer than three; b = training stopped at the 72\,h limit (best checkpoint evaluated, or the seed excluded); s = re-evaluated with the ODE tolerance tightened from $10^{-4}$ to $2.5\times10^{-5}$, after which the numerical check passes; e = one non-finite target of 1{,}023{,}003 excluded; d = one seed diverged during the refit; \dag{} = the numerical check still fails and the value is reported as computed. The main table replaces all cell markers by a single \dag. \emph{Generative models} are not ranked. DSTPP (ub): the model's variational bound (100 diffusion steps), an upper bound on its NLL that is loose by about 2 nats per event by construction; on ComCat, 756 of 21{,}869 targets have waiting times beyond the training range and dominate the mean (marked r; median 4.84). SMASH (a) defines no likelihood: we report its temporal NLL plus the spatial density recovered from its score on a $128\times128$ grid, on 2{,}000 targets per seed (standard error 0.02--0.10). \emph{Outages.} On Chicago and Seattle, the targets that overlap a recording outage are excluded for every model (361 of 50{,}830 and 40 of 24{,}975); Table~\ref{tab:realdata_raw_outage} gives the raw means.}

\paragraph{Rank overview (Figure~\ref{fig:rank_overview}).}\label{sec:rank-notes}
\rev{(a) Rank of each preset with an exact or numerical likelihood within each dataset, by mean test NLL; dark cells: diverged, no value. (b) Mean normalized rank with the median and 90\% interval of a bootstrap over datasets and seeds (2{,}000 replicates). Kendall's $W=0.86$ over the 16 presets ranked on all 13 datasets; Jump-CNF is ranked on 11.}

\paragraph{Calibration (Tables~\ref{tab:calibration_main}--\ref{tab:calib_full_hpd_ks}).}\label{sec:calib-notes}
\rev{Computed from each model's compensator $\Lambda$, without sampling, on the checkpoints and targets of the NLL tables; times in hours. PIT: Kolmogorov--Smirnov statistic of $u_i=1-e^{-\Lambda_i}$ against the uniform law. Timing error: median absolute error of the predictive median waiting time. CRPS$_c$: CRPS of the waiting time censored at $c$, the largest training inter-event time. Coverage: of the central 80\% interval. Spatial HPD-PIT: Kolmogorov--Smirnov statistic of the highest-density-region PIT of the location (n/i: spatially homogeneous preset; ---: not evaluated for the flow decoders). Values are means $\pm$ standard deviations over three seeds; \dag: numerical check not passed, as in the NLL tables; $n$: seeds if fewer than three. Bold: best value per dataset (coverage: closest to 0.80) among cells with three seeds and no \dag. AutoSTPP is evaluated in double precision. SMASH and DSTPP have no compensator: their values are estimated from $K=128$ draws per target on 2{,}000 targets and are not ranked; the same estimator applied to presets with exact samplers reproduces their sampler-free values to within 0.013 (PIT), 0.020 (coverage) and 0.038 (HPD-PIT).}

\paragraph{Next-event prediction (Tables~\ref{tab:nextevent_main}--\ref{tab:nextevent_ranks}).}\label{sec:nextevent-notes}
\rev{Every preset draws $K=128$ samples of the next event for each of 2{,}000 test targets, given the 20 preceding events, and all presets are scored from the draws by the same code. Presets with a likelihood are sampled exactly (waiting time by inverting the cumulative hazard, location from the model's conditional spatial density); SMASH and DSTPP use their own samplers. ES: energy score of $(\min(t,c)/\tau,\ \mathbf{s}/1\,\mathrm{km})$, with $\tau$ the median training inter-event time and $c$ as in the calibration tables. Hit: share of targets whose sample-median forecast is within $\tau$ in time and 1\,km in space; not reported for ComCat and GeoNet, whose spatial cells are 12--26\,km wide. Check: the temporal PIT from the draws differs from the exact one by at most 0.012 over 501 cells. Every preset with at least two seeds is ranked.}

\paragraph{HawkesNest (Table~\ref{tab:suite3_entanglement_test_nll}, Table~\ref{tab:suite3_v2_pit}, Figures~\ref{fig:suite3_budget_curves} and~\ref{fig:post-nll-suite3}).}\label{sec:suite3-notes}
\rev{Excess: model NLL minus the NLL of the true intensity (App.~\ref{app:hawkesnest-generator}) on the common targets of the 20 test windows, mean over three seeds; the seed standard deviation is at most 0.04 (0.10 for the DSTPP bound). The L3$-$L0 column is a paired bootstrap over the test windows (10{,}000 resamples, 95\% interval); bold marks the lowest mean, not a significance claim. Every preset uses the hyperparameters tuned for the suite in the submitted version, unchanged across levels: 50 epochs (AutoSTPP, DeepSTPP, NSMPP), 100 (RMTPP, THP) or 200 (NJSDE, SMASH, DSTPP). Attn-CNF and Jump-CNF were not run because of their training cost. For DSTPP only the variational bound is available, so its true excess is at most the value shown; SMASH has no likelihood. In the learning curves, the excess is computed on the validation events, and the test NLL logged during training agrees with the evaluator to within 0.0054 nats per event. Intensity recovery is the Pearson correlation between the model and the true intensity on a $32\times32$ spatial grid, averaged over 50 time slices per test window.}

\section{Limitations, Compute, and Release Notes}
\label{app:limitations}

\paragraph{Computational cost.}
The benchmark covers model families with very different computational profiles. Factorized baselines and window-based likelihood models are comparatively stable, while continuous-time and flow-based presets require substantially higher wall-clock budgets and are more sensitive to numerical settings. \textsc{Seahorse} records runtime, parameter count, seed, checkpoint, and evaluation metadata in the run artifacts, and failed or unavailable cells are marked explicitly rather than silently removed.

\paragraph{Compute and artifacts.}
Experiments were run mostly in a Slurm-managed HPC environment using containerized Python/PyTorch jobs. \rev{On real data, every neural job used one GPU. The reported seeds ran on A100 (40 or 80\,GB), H100, L40S, A40 or RTX A6000 GPUs of a Slurm cluster with a 72\,h limit per job, or on DGX Spark (GB10) workstations; the statistical fitters ran on CPU. The GPU of every seed is recorded (Figure~\ref{fig:runtime-compute}).} Each fitted run stores \texttt{run\_result.json} metadata including seed, runtime (\texttt{train\_time\_sec}), parameter count (\texttt{n\_params}), selected checkpoint path, effective configuration, normalization statistics, NLL kind and reporting space, test NLL, context counts, and missing-context counts. Post-training evaluation artifacts additionally record the metric profile, evaluation seed, split, device, artifact manifests, elapsed evaluation time where available, checkpoint selection, and checkpoint-test NLL details.

\paragraph{\rev{Incomplete and failed runs.}}
\rev{Of the 741 real-data fits (19 presets $\times$ 13 datasets $\times$ 3 seeds), 715 completed without caveat. Seventeen are reported with a caveat: NSMPP 2 and NJSDE 3 (numerical check not passed), Attn-CNF 6 (3 re-evaluated with a tightened solver, 1 check not passed, 2 stopped at 72\,h), and Jump-CNF 6 (3 re-evaluated, 3 checks not passed). Two Attn-CNF seeds were excluded after the 72\,h limit without a usable checkpoint, and seven Jump-CNF seeds diverged: six in every tuning trial on Citi Bike (h) and NYC~311, one during the refit on ComCat. Table~\ref{tab:real_data_full_nll} marks the affected cells, Table~\ref{tab:preset_audit} counts them per preset, and Figure~\ref{fig:runtime-reliability} shows every seed.}

\paragraph{Metric availability.}
Not all diagnostics apply to all model families. Surface-recovery metrics require known ground-truth intensity and therefore apply only to controlled synthetic suites. Sample-based models such as SMASH and DSTPP do not generally expose direct pointwise intensities, so intensity-surface diagnostics require an explicit approximation or are marked unavailable.

\paragraph{Benchmark scope.}
The current experiments focus on fixed train/validation/test splits, raw-space likelihood reporting, post-NLL diagnostics, and controlled entanglement stress tests. The framework is designed to support additional datasets, metrics, and model presets, but the present paper does not claim exhaustive coverage of all STPP modeling regimes. In particular, covariate-conditioned STPPs \rev{are} left for future extensions.

\paragraph{Release.}
The release includes the \textsc{Seahorse} codebase, preset configurations, dataset schema documentation, benchmark scripts, and instructions for \rev{reproducing the main results}. \rev{The code is available at \repo{} and as the Python package \texttt{seahorse-stpp} (version 1.0.1). The datasets are available at \href{https://huggingface.co/seahorse-stpp}{\texttt{huggingface.co/seahorse-stpp}}; for Citi Bike and Seattle~911 we provide scripts that rebuild the data from the original source instead of redistributing it.}

\end{document}